\documentclass[preprint,12pt]{elsarticle}

\usepackage{listings}
\usepackage[bookmarks=false]{hyperref}
\usepackage{orcidlink}

\usepackage[many]{tcolorbox}
 \newtcolorbox{mydoublebox}[1][]{enhanced,
  colframe=blue!75!black,
  width=\linewidth,
}

\usepackage[indLines=false]{algpseudocodex}%
\algrenewcommand\algorithmicrequire{\textbf{Input:}}

 \mathchardef\mhyphen="2D

 \usepackage{graphicx}
\usepackage{multirow}
\usepackage{subcaption}
\usepackage{amsmath,amsfonts,amssymb}
\newtheorem{definition}{Definition}
\usepackage{MnSymbol}
\usepackage{algorithm}%
\usepackage{tikz}
\usetikzlibrary{automata,positioning}
\usepackage{forest}
\usepackage{caption}
\usepackage{subcaption}
\usepackage[table,xcdraw]{xcolor}

\usepackage{lscape}
\usepackage{rotating}
\usepackage{multicol}

\usepackage[indLines=false]{algpseudocodex}%
\algrenewcommand\algorithmicrequire{\textbf{Input:}}

\usepackage{array}

\definecolor{Ronecolor}{RGB}{0, 0, 255} 
\definecolor{Rtwocolor}{RGB}{0, 191, 255} 
\definecolor{Rthreecolor}{RGB}{30, 144, 255} 
\definecolor{Rfourcolor}{RGB}{65, 105, 225} 

 \mathchardef\mhyphen="2D

\begin{document}

\begin{frontmatter}



\title{Integrating Domain Knowledge into Process Discovery Using Large Language Models}



\author[aff1]{Ali Norouzifar\orcidlink{0000-0002-1929-9992}}
\ead{ali.norouzifar@pads.rwth-aachen.de}

\author[aff1,aff2]{Humam Kourani\orcidlink{0000-0003-2375-2152}}
\ead{humam.kourani@fit.fraunhofer.de}

\author[aff3]{Marcus Dees\orcidlink{0000-0002-6555-320X}}
\ead{marcus.dees@uwv.nl}

\author[aff1]{Wil van der Aalst\orcidlink{0000-0002-0955-6940}}
\ead{wvdaalst@pads.rwth-aachen.de}

\affiliation[aff1]{organization={RWTH Aachen University}, 
            city={Aachen},
            country={Germany}}

\affiliation[aff2]{organization={Fraunhofer FIT}, 
            city={Sankt Augustin},
            country={Germany}}

\affiliation[aff3]{organization={UWV Employee Insurance Agency}, 
            city={Amsterdam},
            country={Netherlands}}
            
\begin{abstract}
Process discovery aims to derive process models from event logs, providing insights into operational behavior and forming a foundation for conformance checking and process improvement. However, models derived solely from event data may not accurately reflect the real process, as event logs are often incomplete or affected by noise, and domain knowledge, an important complementary resource, is typically disregarded. As a result, the discovered models may lack reliability for downstream tasks. We propose an interactive framework that incorporates domain knowledge, expressed in natural language, into the process discovery pipeline using Large Language Models (LLMs). Our approach leverages LLMs to extract declarative rules from textual descriptions provided by domain experts. These rules are used to guide the IMr discovery algorithm, which recursively constructs process models by combining insights from both the event log and the extracted rules, helping to avoid problematic process structures that contradict domain knowledge. The framework coordinates interactions among the LLM, domain experts, and a set of backend services. We present a fully implemented tool that supports this workflow and conduct an extensive evaluation of multiple LLMs and prompt engineering strategies. Our empirical study includes a case study based on a real-life event log with the involvement of domain experts, who assessed the usability and effectiveness of the framework.
\end{abstract}


\begin{highlights}
\item Proposes a novel framework for integrating domain knowledge into process discovery using Large Language Models (LLMs).
\item Demonstrates the transformation of textual process descriptions into declarative rules via LLMs to guide model construction.
\item Supports iterative refinement of process models through interactive feedback from domain experts and LLM-generated rule updates.
\item Provides a comprehensive evaluation of multiple LLMs and prompting strategies based on ground truth constraints.
\item Implements a user-friendly tool enabling end-to-end discovery, expert interaction, and visualization of process models and rule statistics.
\end{highlights}


\end{frontmatter}



\section{Introduction}
Event data recorded by information systems provides a valuable foundation for process mining, enabling organizations to gain insights into their operations and identify opportunities for improvement. In this context, a variety of automated techniques are applied to discover descriptive models that represent and explain observed process behavior~\cite{DBLP:journals/tkde/AugustoCDRMMMS19}. To evaluate the quality of these discovered models with respect to a given event log, conformance checking metrics such as \emph{fitness}, \emph{precision}, \emph{generalization}, and \emph{simplicity} are commonly used~\cite{CarmonaDSW18}. These models serve as a starting point for various types of analysis, including the detection of deviations, performance evaluation, and process enhancement.

\begin{figure}[htb]
    \centering
    \includegraphics[width=0.6\linewidth]{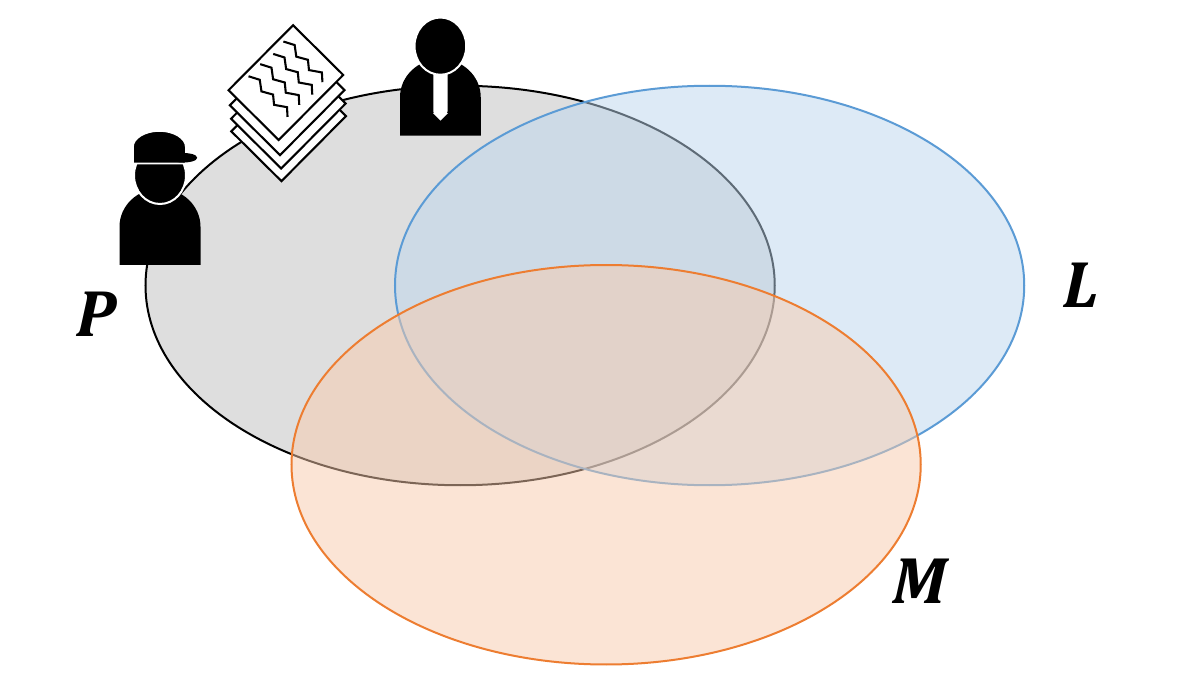}
    \caption{Venn diagram illustrating the relationship between the event log ($L$), the discovered model ($M$), and the actual process ($P$).}

    \label{fig:van_diag}
\end{figure}

 As illustrated by the Venn diagram in Fig.~\ref{fig:van_diag}, the central objective of process discovery is to derive a model (represented with $M$) that reflects the behavior recorded in the event log (represented with $L$). However, the broader goal is to approximate the \emph{actual} underlying process (represented with $P$), which is typically unknown. This poses a challenge: the event log may not fully capture the real process, i.e., there may be parts of the process in $P - L$ that are never observed, and it may include traces that deviate from the expected behavior due to noise or exceptional circumstances, i.e., elements in $L - P$. As a result, evaluating the discovered model solely based on the event log can be misleading, since the log represents only a partial and potentially imperfect view of the actual process, particularly when these models are intended for conformance checking and process improvement.

In addition to event data extracted from information systems, organizations often have access to valuable resources such as domain experts, process documentation, and other forms of \emph{domain knowledge}~\cite{DBLP:journals/cii/SchusterZA22}. While this knowledge cannot be directly leveraged by traditional process discovery techniques, it provides complementary insights into how the actual process operates. However, such resources typically remain underutilized during model discovery. Incorporating domain knowledge into the process discovery task presents several challenges. For example, domain experts possess deep understanding of their processes but usually express their knowledge in natural language, which is difficult to formalize. Similarly, textual process documentation, despite its richness, poses integration challenges due to its unstructured nature. In this work, we address these limitations by enabling the direct incorporation of domain knowledge into process discovery using Large Language Models (LLMs). The recent emergence of powerful LLMs, such as GPT-4~\cite{DBLP:journals/corr/abs-2303-08774} and Gemini~\cite{DBLP:journals/corr/abs-2312-11805}, which demonstrate strong capabilities in natural language understanding and reasoning over textual inputs, offers promising opportunities for advancing tasks related to text interpretation.

This paper builds upon the original idea of incorporating domain knowledge, transformed into formal rules via an LLM within a specific process discovery technique, namely the IMr framework~\cite{DBLP:conf/bpm/NorouzifarKDA24}. IMr is an inductive mining-based approach that recursively identifies the process structure that best represents the underlying process, using both an event log and a set of input rules~\cite{DBLP:conf/rcis/NorouzifarDA24}. The framework systematically explores alternative process structures and leverages the provided rules to prune the search space, thereby eliminating suboptimal candidates. While the notion of rules can be broadly interpreted, this work adopts the Declare rule specification language as a representative formalism~\cite{DBLP:conf/caise/MaggiBA12}. Declarative rules are particularly suitable for this purpose due to their alignment with human reasoning and their well-defined logical semantics, which are supported by a rich body of literature. Nonetheless, expecting end users to manually define such rules is often unrealistic, motivating the use of LLMs to bridge this gap.

\begin{figure}[htb]
    \centering
    \includegraphics[width=0.5\linewidth]{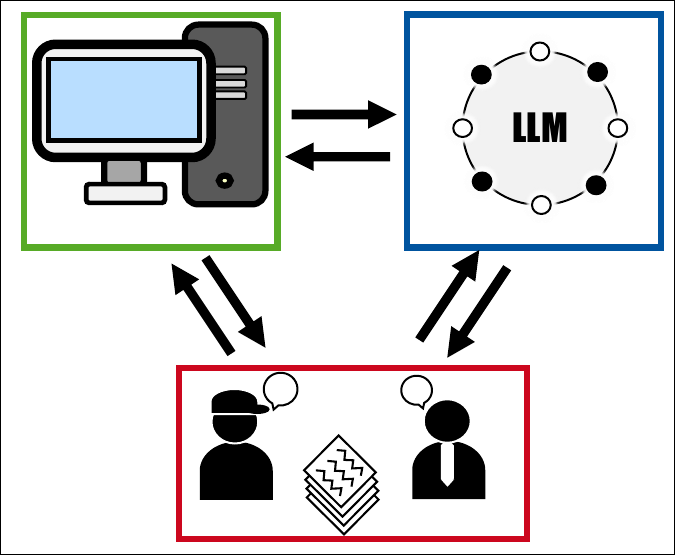}
    \caption{Overview of the interactive framework, showing how the LLM, domain expert, and backend services collaborate to support rule extraction and process discovery.}
    \label{fig:interactions}
\end{figure}

Our proposed method goes beyond a simple chatbot by integrating multiple interacting components. Figure~\ref{fig:interactions} illustrates the interconnections among the key entities in the framework: domain knowledge (represented by the red box), the large language model (blue box), and the system components (green boxes), which include several predefined services responsible for managing communication and executing algorithms. Domain knowledge can be injected into the discovery process either prior to the initial model construction or after a preliminary model has been discovered and reviewed by domain experts. In both scenarios, an LLM, configured through prompt engineering, acts as an intelligent assistant that interfaces with available resources. The LLM engages in dialogues with the domain expert to acquire process descriptions and to ask clarification questions when ambiguities arise. The services autonomously handle tasks such as generating task descriptions, managing information about event logs, validating LLM outputs, and extracting rules from them, without requiring direct intervention from the domain expert. Meanwhile, the domain expert can interact with the system to perform actions such as initiating process discovery, reviewing statistical summaries, and selecting plausible rules based on domain knowledge.

The main contributions of this paper, in comparison to our previous work~\cite{DBLP:conf/bpm/NorouzifarKDA24}, are summarized as follows:

\begin{itemize}
    \item \textbf{Framework Redesign for Transparency and Automation:}  
    We redesigned the framework to incorporate several dedicated backend services responsible for executing predefined algorithms. These services enhance automation and reliability, while also promoting more active involvement of the domain expert. This design reduces the risk of opaque or unreliable rule communication, thereby improving the transparency and explainability of the LLM-generated rules.

    \item \textbf{Tool Development and Interactive User Interface:}  
    We developed a fully functional tool that implements the proposed framework. It features a user-friendly web interface that allows users to run the entire pipeline by selecting an LLM model and providing an API key. The tool supports interactive exploration of results and enables users to input process descriptions at different stages. This interactivity facilitates iterative refinement of the extracted rules and discovered models.

    \item \textbf{Extended Evaluation with Multiple LLMs and Prompting Strategies:}  
    We conducted a comprehensive quantitative evaluation involving several LLMs and prompt engineering strategies. The evaluation assesses the models' ability to extract declarative rules from natural language descriptions, comparing their outputs against a human-labeled ground truth.

    \item \textbf{Evaluation Metrics for Accuracy and Robustness:}  
    We designed and measured several evaluation metrics to compare different configurations. These metrics include recall, precision, the frequency of error-handling cycles, and the failure rate when no valid rule set can be generated after multiple correction attempts.

    \item \textbf{Assessment of LLM Behavior Under Ambiguity:}  
    Since clarification dialogues are expected in our method when input is ambiguous, we evaluated how different LLMs handle such cases. The analysis focuses on whether the models appropriately engage in clarification and how effectively they recover from incomplete or unclear input.
\end{itemize}

Section~\ref{sec:related_work} reviews related work on automated process discovery, the use of domain knowledge, and recent applications of large language models in process mining. Section~\ref{sec:pre} introduces the necessary preliminaries and presents the IMr technique, which discovers process models based on event logs and declarative rules. In Section~\ref{sec:framework}, we describe our proposed method for extracting rules from textual inputs, along with the coordination between different system components to ensure reliable results. Section~\ref{sec:tool} presents the web-based tool developed to support the framework and facilitate user interaction. Section~\ref{sec:evaluation} reports the results of a comprehensive evaluation, including a real-world case study demonstrating the practical applicability of the framework. Finally, Section~\ref{sec:conclusion} summarizes the proposed approach and key findings, and concludes the paper.

\section{Related Work}
\label{sec:related_work}
Traditional process discovery methods primarily rely on event data to construct process models~\cite{DBLP:journals/tkde/AugustoCDRMMMS19}. The overarching goal is to generate models that not only fit the observed behavior well (i.e., exhibit high fitness), but also restrict behavior not present in the log (i.e., ensure high precision), allow for plausible but unseen behavior (i.e., achieve good generalization), and maintain a level of simplicity that makes them easy to interpret and understand~\cite{CarmonaDSW18}.
 
Event data is not the only resource available for process discovery. In many real-world scenarios, domain knowledge, such as expert insights and process documentation, can be leveraged as an additional input across various process mining tasks~\cite{DBLP:journals/cii/SchusterZA22}. This knowledge can be utilized prior to discovery, for instance, to filter the event log~\cite{DBLP:conf/caise/EckLLA15}. Some discovery techniques, such as the one proposed in~\cite{DBLP:conf/rcis/NorouzifarDA24}, are designed to incorporate domain knowledge in the form of declarative rules alongside the event log, thereby guiding the discovery process toward models that better reflect expert expectations. The approach presented in~\cite{DBLP:conf/er/DixitVBA18} introduces an interactive framework in which users provide low-level input by manually placing transitions and places, supported by visual guidance. The method in~\cite{DBLP:journals/softx/SchusterZA23} begins with an initial model discovered from a user-selected subset of variants and iteratively refines it by incorporating additional variants to incrementally update the model.

Declarative constraints are widely used in the literature to represent process models that characterize the allowed behavior in the business processes~\cite{DBLP:books/sp/22/CiccioM22}. Despite their strong formal semantics, these constraints are often interpretable in natural language, making them accessible to domain experts~\cite{DBLP:conf/caise/AaCLR19}.  
Building on the IMr discovery technique proposed in~\cite{DBLP:conf/rcis/NorouzifarDA24}, the framework proposed in~\cite{DBLP:conf/bpm/NorouzifarKDA24} leverages large language models to assist domain experts in translating textual process descriptions into declarative constraints. The extracted rules are then integrated into the discovery approach, enabling the generation of imperative process models that align more closely with domain knowledge by combining both the event log and the derived constraints.

Some studies focus on repairing process models after the discovery phase, aiming primarily to enhance alignment between the discovered models and the corresponding event logs, rather than explicitly incorporating domain knowledge~\cite{DBLP:journals/tosem/PolyvyanyyAHW17,DBLP:journals/is/FahlandA15}. The main objective of such approaches is to improve the conformance between the model and the log. In contrast, our work not only integrates domain knowledge expressed in natural language prior to the discovery phase, but also enables the incorporation of textual feedback from domain experts. This allows for refining the discovered process models in accordance with expert expectations and domain-specific requirements.

The paper~\cite{DBLP:journals/eis/WoenselM24} offers a comprehensive survey of NLP techniques used for extracting business rules from textual sources. The transformation of natural language process descriptions into declarative constraints was explored in earlier works~\cite{DBLP:conf/caise/AaCLR19,DBLP:conf/bpm/AaBMN20}. Later the approaches like~\cite{DBLP:conf/bpm/GrohsAER23} showed that the extraction of such rules with LLM is more effective by comparing the results of traditional NLP methods with LLM extracted rules.
More recently,~\cite{DBLP:conf/coopis/NeubergerAJ23} proposed a fully data-driven pipeline based on gradient boosting to extract business rules from natural language. Similar to declarative constraints, these rules capture logical relationships between activity instances derived from text.

The paper~\cite{DBLP:conf/coling/AaCLMP18} explores how natural language processing techniques support various levels of business process management, including the transformation of textual descriptions into formal models, verification of model completeness, and conversational process guidance. With the rise of LLMs, NLP has become more scalable and effective. LLMs have been applied to diverse process mining tasks. The study in~\cite{DBLP:conf/bpm/VidgofBM23} investigates their integration across the BPM lifecycle, from identification and discovery to analysis, redesign, implementation, and monitoring. It showcases application scenarios such as extracting processes from unstructured documentation, generating models from communication logs, and facilitating conversational interfaces for interactive process analysis.

From a system-level perspective, LLMs are increasingly integrated into conversational BPM systems to improve transparency and user engagement. The paper~\cite{DBLP:conf/rcis/CascianiBCM24} shows how LLM-powered chatbots can transform traditional BPM systems into AI-augmented ones that proactively assist users, provide recommendations, and enhance explainability. 
The paper~\cite{DBLP:conf/icws/BeheshtiYSBCDMZX23} proposes ProcessGPT, a generative AI-based framework designed to enhance business process management. It leverages large language models to support both automation and augmentation in data-centric and knowledge-intensive processes. 
The paper~\cite{DBLP:conf/worldcist/AyadA24} investigates how large language models can support and improve the semantic quality of business process models. 


The use of LLMs for process modeling has received increasing attention, primarily focusing on the extraction of formal models from natural language texts. The advent of LLMs has substantially advanced efforts to bridge the gap between unstructured domain knowledge and formal process representations. The study~\cite{DBLP:conf/bpm/KlievtsovaBKMR23} surveys methods for deriving process models using chatbots and examines how LLM-powered conversational services can assist in the creation and refinement of models through natural language interaction with domain experts. Several other works demonstrate the capability of LLMs to transform textual descriptions into process artifacts, including process models~\cite{DBLP:conf/icsoc/NivonS24,DBLP:conf/ijcai/KouraniB0A24,DBLP:conf/models/IvanchikjSP20}. In contrast to these approaches, our method retains the event log as the primary source of information and augments the discovery process by integrating textual process knowledge.

The study in~\cite{DBLP:conf/bpm/Berti0A23} surveys various prompt engineering strategies tailored for process mining applications. Recent research also investigates prompt engineering and automated prompt correction to enhance the robustness of LLMs. The paper~\cite{DBLP:conf/bpmds/BuschRSL23} highlights prompt engineering as a promising approach for applying LLMs in BPM, emphasizing the use of task-specific prompts to guide pre-trained models without requiring fine-tuning. A foundational contribution in this context is GPT-3~\cite{DBLP:conf/nips/BrownMRSKDNSSAA20}, which demonstrated that LLMs can perform diverse NLP tasks in few-shot, one-shot, or even zero-shot settings by relying solely on prompt-based instructions. The paper ~\cite{DBLP:conf/iclr/ZhouMHPPCB23} proposes the automatic prompt engineer method, where one LLM generates and scores candidate prompts to autonomously select the most effective one, achieving performance comparable to human experts. In a related direction,~\cite{DBLP:journals/dke/FahlandFLSS25} explores the use of LLMs for generating user-centered explanations of business processes. Their situation-aware explainability framework combines multiple services to tailor explanations to the user's perspective and context.

The paper~\cite{DBLP:journals/tacl/PanSXNWW24} provides a comprehensive survey of self-correction techniques for LLMs. It categorizes existing methods into three main stages, training-time, generation-time, and post-hoc correction, and examines how feedback (either internal/self-generated or external) can be leveraged to address issues such as hallucination.
The paper~\cite{DBLP:conf/iclr/0009CMZYSZ24} critically evaluates the self-correction capabilities of LLMs, particularly in reasoning tasks. It introduces the concept of intrinsic self-correction, where models attempt to improve their own responses without any external feedback.
Techniques such as chain-of-thought prompting, retrieval-augmented generation, and feedback loops are used to enhance model reliability and interpretability in BPM contexts.


\section{Background}
\label{sec:pre}
This section introduces the necessary preliminaries and outlines the underlying process discovery technique that forms the foundation of our framework.
\subsection{Preliminaries}
Given a set \( B \), the notation \( B^* \) denotes the set of all finite sequences that can be formed from elements of \( B \). The notation \( \mathcal{P}(B) \) refers to the power set of \( B \), i.e., the set of all subsets of \( B \). The set $\mathcal{A}$ represents the universe of activities.

\begin{definition}[Event Log]
A \emph{trace} $\sigma = \langle a_1, a_2, \ldots, a_n \rangle \in {\mathcal{A}}^*$ is a finite sequence of activities, where each activity occurrence corresponds to an \emph{event}. An \emph{event log} $L \in \mathcal{B}({\mathcal{A}}^*)$ is a multiset of traces, capturing multiple executions of the process. We denote the universe of all possible event logs by $\mathcal{L}$.
\end{definition}

Table~\ref{tab:event_log} presents an example event log containing 13 events distributed across 5 traces. The individual traces are defined as follows:
\begin{align*}
    \sigma_1 &= \langle \texttt{A-created}, \texttt{Doc-checked}, \texttt{Hist-checked}, \texttt{A-rejected} \rangle,\\ 
    \sigma_2 &= \langle \texttt{A-created}, \texttt{Hist-checked}, \texttt{Doc-checked}, \texttt{A-accepted} \rangle, \\
    \sigma_3 &= \langle \texttt{A-created}, \texttt{A-canceled} \rangle, \sigma_4 = \langle \texttt{A-created}, \texttt{A-canceled} \rangle,\\
    \sigma_5 &= \langle \texttt{A-canceled} \rangle
\end{align*}

\begin{table}[tb]
\caption{Example event log extracted from a claim-handling process.}
\label{tab:event_log}
\centering
\begin{tabular}{|c|c|c|c|}
\hline
\textbf{Event} & \textbf{Case ID} & \textbf{Activity}       \\ \hline
$e_1$ & \textit{1}              & \texttt{A-created}          \\ \hline
$e_2$ & \textit{2}              & \texttt{A-created}       \\ \hline
$e_3$ & \textit{1}              & \texttt{Doc-checked}       \\ \hline
$e_4$ & \textit{2}              & \texttt{Hist-checked}        \\ \hline
$e_5$ & \textit{3}              & \texttt{A-created}        \\ \hline
$e_6$ & \textit{2}              & \texttt{Doc-checked}           \\ \hline
$e_7$ & \textit{1}              & \texttt{Hist-checked}         \\ \hline
$e_8$ & \textit{2}                & \texttt{A-accepted}       \\ \hline
$e_9$ & \textit{3}                & \texttt{A-canceled}        \\ \hline
$e_{10}$ & \textit{1}              & \texttt{A-rejected}      \\ \hline
$e_{11}$ & \textit{4}                & \texttt{A-created}  \\ \hline
$e_{12}$ & \textit{4}                & \texttt{A-canceled} 
\\ \hline
$e_{13}$ & \textit{5}                & \texttt{A-canceled}\\ \hline

\end{tabular}
\end{table}

The corresponding event log is represented as
\begin{align*}
    L_1= [& \langle \texttt{A-created}, \texttt{Doc-checked}, \texttt{Hist-checked}, \texttt{A-rejected} \rangle,\\ 
     & \langle \texttt{A-created}, \texttt{Hist-checked}, \texttt{Doc-checked}, \texttt{A-accepted} \rangle, \\
    &\langle \texttt{A-created}, \texttt{A-canceled} \rangle^2, \langle \texttt{A-canceled} \rangle]
\end{align*}

Declarative rules are instantiated from predefined rule templates, each specifying a particular type of behavioral constraint between activities. 

\begin{definition}[Rules]
\label{Def:rules}
Let $\mathcal{T}$ denote a non-empty set of templates, where each template is a relation of the form $d(x_1, \dots, x_m) \in \mathcal{T}$, defined over variables $x_1, \dots, x_m$, with $m \in \mathbb{N}$ indicating the arity of the template $d$. Given a set of activities $a_1, \dots, a_m \in \mathcal{A}$, a \emph{rule} is an instantiation of the template $d$, written as $d(a_1, \dots, a_m)$, where each variable is bound to a corresponding activity. We denote the universe of all possible rules by $\mathcal{R}$.
\end{definition}

In this paper, we focus on a subset of commonly used templates from the declarative process modeling literature. These templates capture various behavioral relations and serve as the basis for formulating rules that describe expected process behavior. 
The universe of templates $\mathcal{T}$ considered in this work include the following:
\begin{itemize}
    \item $AtMost1(x_1)$: $x_1$ occurs at most once.
    \item $AtLeast1(x_1)$: $x_1$ occurs at least once.
    \item $Response(x_1,x_2)$: If $x_1$ occurs, then $x_2$ occurs after $a$.
    \item $Precedence(x_1,x_2)$: $x_2$ occurs only if preceded by $x_1$.
    \item $RespondedExistence(x_1,x_2)$: If $x_1$ occurs in the trace, then $x_2$ occurs as well.
    \item $CoExistence(x_1,x_2)$: $x_1$ and $x_2$ occur together.
    \item $NotCoExistence(x_1,x_2)$: $x_1$ and $x_2$ never occur together.
    \item $NotSuccession(x_1,x_2)$: $x_2$ cannot occur after $x_1$. 
\end{itemize}

For example, based on the predefined templates and the activities in the event log shown in Table~\ref{tab:event_log}, we can define the following rules: 
\begin{align*}
r_1 &= \mathit{Precedence}(\texttt{A-created}, \texttt{A-canceled}),\\
r_2 &= \mathit{NotSuccession}(\texttt{Doc-checked}, \texttt{Hist-checked})\\
r_3 &= \mathit{Response}(\texttt{A-created}, \texttt{Hist-checked})
\end{align*}

Each rule imposes constraints on the execution of traces based on the semantics of its underlying template. For instance,  \( \mathit{Precedence}(\texttt{A-created},\) \(\texttt{A-canceled}) \) is violated by any trace in which \texttt{A-canceled} occurs without a preceding occurrence of \texttt{A-created}. We refer to the set of all traces that satisfy a given rule as the \emph{language} of the rule.

\begin{definition}[Language of a Rule]
Let $r \in \mathcal{R}$ be a rule. A trace $\sigma \in \mathcal{A}^{*}$ is considered to satisfy the rule $r$ unless it explicitly violates the constraint, according to the semantics of the underlying template. We denote such a violation by $\sigma \nvDash r$. The \emph{language} of a rule, denoted by $\phi(r)$, is defined as the set of all traces that satisfy the rule:
\[
\phi(r) = \{\sigma \in \mathcal{A}^{*} \mid \sigma \vDash r\}.
\]
\end{definition}

Given the example event log and the defined rules, we can analyze the satisfaction and violation of constraints as follows:
\begin{itemize}
    \item $\sigma_5 \nvDash d_1$ because the application is canceled (event \texttt{A-canceled}) without a preceding occurrence of \texttt{A-created}, which is expected to indicate the creation of the application. Hence, $\sigma_5 \notin \phi(d_1)$.
    
    \item $\sigma_1 \nvDash d_2$ since the event \texttt{Hist-checked} occurs after \texttt{Doc-checked}, violating the expected order. Therefore, $\sigma_1 \notin \phi(d_2)$.
    
    \item $\sigma_2 \vDash d_3$ because \texttt{A-created} appears in the trace, and \texttt{Hist-checked} occurs subsequently, which complies with the rule. Thus, $\sigma_2 \in \phi(d_3)$.
\end{itemize}

Since declarative constraints are expressed as rule-based constructs, their quality can be evaluated using standard metrics such as \emph{support} and \emph{confidence}. In our framework, these metrics are computed and reported to the domain expert, providing statistical insights that assist in assessing the relevance and reliability of the extracted rules with respect to the given event log.

\begin{definition}[Support and Confidence]
\label{Def:declare_evaluation}
Let \( L \in \mathcal{L} \) be an event log. For a given rule \( r \in \mathcal{R} \), we define the following quantities:
\begin{itemize}
    \item \( \#_{\mathit{activated}}(L, r) \): the number of traces the constraint \( r \) is activated in \( L \),
    \item \( \#_{\mathit{satisfied}}(L, r) \): the number of traces \( r \) is satisfied in \( L \),
    \item \( \#_{\mathit{traces}}(L) \): the total number of traces in the log \( L \).
\end{itemize}
Based on these quantities, the following metrics are defined:
\begin{itemize}
    \item \textbf{Support:} The proportion of traces satisfying the constraint relative to the total number of traces in the log:
    \[
    \mathit{Support}(L, r) = \frac{\#_{\mathit{satisfied}}(L, r)}{\#_{\mathit{traces}}(L)}.
    \]
    \item \textbf{Confidence:} The proportion of traces satisfying the constraint among all traces with an activation of the constraint. To prevent division by zero, the denominator is lower-bounded by 1:
    \[
    \mathit{Confidence}(L, r) = \frac{\#_{\mathit{satisfied}}(L, r)}{\max(1, \#_{\mathit{activated}}(L, r))}.
    \]
\end{itemize}
\end{definition}

Considering the event log $L_1$ and the example rules, 
\begin{itemize}
    \item $support(L_1,r_1) = \frac{2}{5}$, $confidence(L_1,r_1)=\frac{2}{3}$
    \item $support(L_1,r_2) =\frac{1}{5}$, $confidence(L_1,r_2)=\frac{1}{2}$
    \item $support(L_1,r_3) =\frac{2}{5}$, $confidence(L_1,r_3)=\frac{2}{5}$
\end{itemize}

\subsection{IMr Process Discovery}
\label{sec:IMr_intro}
\begin{figure}[htb]
    \centering
    \includegraphics[width=1\linewidth]{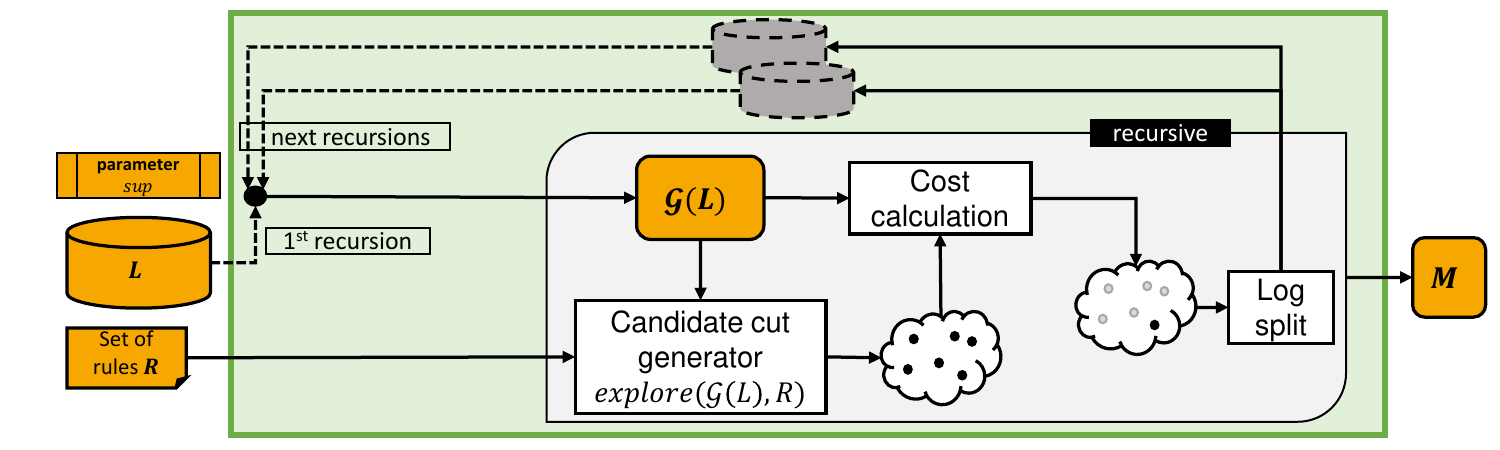}
    \caption{An overview of the IMr process discovery framework~\cite{DBLP:conf/rcis/NorouzifarDA24}.}
    \label{IMr+LLM_framework}
\end{figure}

Figure~\ref{IMr+LLM_framework} provides an overview of the IMr framework, and Algorithm~\ref{IM-rules} illustrates how its individual components interact to recursively discover a process model, as introduced in~\cite{DBLP:conf/rcis/NorouzifarDA24}. The IMr framework adopts a divide-and-conquer strategy for process discovery, recursively identifying process structures by partitioning the behavior observed in the event log.
The algorithm begins with the following inputs:
\begin{itemize}
    \item an event log \(L \in \mathcal{L}\), capturing observed process executions,
    \item a predefined set of declarative rules \(R \in \mathcal{R}\), which guide and constrain the discovery process, and
    \item a parameter \(sup \in [0,1]\),  which regulates the penalization of missing behavior during model discovery.
\end{itemize}

This framework assumes that the set of rules $R$ is provided as input.
However, when domain knowledge is available in textual form, manually extracting these rules is not straightforward. In the following section, we describe how such rules can be automatically derived from natural language inputs.

\begin{algorithm}[H]
            \caption{\small $IMr$ algorithm}
            \label{IM-rules}
            \begin{algorithmic}[1]
            \Function{$IMr$}{$L,sup,R$}
                 \LComment{$L {\in} {\mathcal{L}}$ is an event log, $sup \in [0,1]$ is a process discovery parameter and $R$ is the set of rules.}
                \State $base = checkBaseCase(L,sup)$
                \If {$checkBaseCase$ successful}
                    \State \Return $base$
                \EndIf
                \State $C = explore(\mathcal{G}(L),R)$
                \State $(\oplus,\Sigma_1,\Sigma_2)=\underset{c \in C}{\arg\min} \lbrace ov\_cost_{\mathcal{G}(L)}(c, sup) \rbrace$
                \State $L_1, L_2 = SPLIT(L,(\oplus,\Sigma_1,\Sigma_2))$
                \State \Return $\oplus(IMr(L_1,sup,R),IMr(L_2, sup,R))$
                \EndFunction
            \end{algorithmic}
        \end{algorithm}

The first step of IMr is to search for base cases, i.e., check whether the event log consists of only a single activity. In such cases, the process model consists of just that one activity. If no base cases are found, IMr employs the directly-follows graph as an intermediate representation, formally defined below, to determine a suitable process structure at each recursion step.

\begin{definition} [Directly-Follows Graph]
Let $L \in \mathcal{L}$ be an event log. The DFG of this event log denoted as \(\mathcal{G}(L) = (\Sigma, E)\), is constructed, where:
\begin{itemize}
    \item \(\Sigma = \{a \mid a \in \sigma, \, \sigma \in L \}\) represents the set of activities, and 
    \item \(E = [(\sigma(i), \sigma(i+1)) \mid \sigma \in L', \, 1 \leq i < |\sigma|]\), with \(L' = [\langle \triangleright \rangle . \sigma . \langle \square \rangle \mid \sigma \in L]\), denotes the multiset of edges capturing directly-follows relationships. Here, artificial start (\(\triangleright\)) and end (\(\square\)) nodes are added, ensuring completeness in the graph representation.
\end{itemize}
\end{definition}

At each recursion step, the algorithm extracts a directly-follows graph from the event log. Its task is to partition the set of activities into two subsets, \( \Sigma_1 \) and \( \Sigma_2 \), based on a selected cut operator, i.e., sequence, exclusive choice, concurrent, or loop type, that defines the behavioral relation between the activities in these subsets.

\begin{definition}[Binary Cut]
Let  $L {\in} {\mathcal{L}}$ be an event log. $\mathcal{G}(L){=}(\Sigma, E)$ is the corresponding DFG. A binary cut $(\oplus,\Sigma_1,\Sigma_2)$ divides $\Sigma$ into two partitions, such that $\Sigma_1 {\cup} \Sigma_2 {=} \Sigma$, $\Sigma_1  {\cap} \Sigma_2 {=} \emptyset$, and $\oplus {\in} \lbrace \rightarrow,\times, \wedge, \circlearrowleft \rbrace$ is a cut type operator. $\mathcal{C}_{\Sigma}$ is the universe of binary cuts we can generate with the set of nodes $\Sigma$.
\end{definition}

Given a set of rules \( R \), the algorithm leverages these rules to prune the space of all possible cuts \( \mathcal{C}_{\Sigma} \), retaining only those that do not violate any of the specified constraints. Since there are numerous ways to partition the set of activities using different cut operators, the space of potential cuts can be large. The function \( \mathit{explore}(\mathcal{G}(L), R) \) is responsible for identifying the subset of candidate cuts that are consistent with the rules in \( R \).

Constraint violation in a cut is a key concept that guides the process discovery algorithm by eliminating candidate cuts which, regardless of the chosen process structure in the next recursion, result in traces that violate at least one rule in \( R \). We formally define the notion of constraint violation in the context of cuts as follows.

\begin{definition}[Constraint Violation]
Let \( L \in \mathcal{L} \) be an event log, and let \( \mathcal{G}(L) = (\Sigma, E) \) denote the directly-follows graph extracted from \( L \). Let \( R \) be a set of rules.  A cut \( c = (\oplus, \Sigma_1, \Sigma_2) \in \mathit{explore}(\mathcal{G}(L), R) \) is said to \emph{violate} a constraint \( r \in R \), denoted by \( c \nvDash r \), if the following condition holds:

for every process tree \( M \in \mathcal{M}_c \), where
\[
\mathcal{M}_c = \{\, \oplus(M_1, M_2) \mid M_1 \in \mathcal{M}_{\Sigma_1},\; M_2 \in \mathcal{M}_{\Sigma_2} \,\},
\]
there exists a trace \( \sigma \in \phi(M) \) such that \( \sigma \nvDash r \). In other words, none of the process models induced by the cut \( c \) can satisfy the constraint \( r \) for all their traces. Here, \( \mathcal{M}_{\Sigma_1} \) and \( \mathcal{M}_{\Sigma_2} \) represent the sets of all possible process trees over activity sets \( \Sigma_1 \) and \( \Sigma_2 \), respectively.
\end{definition}

For example, consider the cut 
\begin{align*}
c = (&\rightarrow, \{\texttt{A-created}, \texttt{Doc-checked}\}, \{\texttt{Hist-checked}, \texttt{A-accepted}, \texttt{A-rejected},\\ &\texttt{A-canceled}\})
\end{align*}
and the rule 
\[
r_2 = \mathit{NotSuccession}(\texttt{Doc-checked}, \texttt{Hist-checked}).
\]
We have \( c \nvDash r_2 \) because all process models that can be generated from this cut allow for at least one trace where \texttt{Doc-checked} occurs before \texttt{Hist-checked}, thereby violating \( r_2 \).

Cost functions are used to evaluate the quality of each candidate cut by considering the number of deviating edges and estimating missing edges with respect to a given threshold \( \mathit{sup} \). For example, in the case of a sequence cut, any edge from \( \Sigma_2 \) to \( \Sigma_1 \) is considered a deviation and contributes to the cost. Additionally, edges from \( \Sigma_1 \) to \( \Sigma_2 \) are expected; their absence results in a penalty based on the \( \mathit{sup} \) parameter. A higher \( \mathit{sup} \) value enforces stricter requirements on the presence of such expected edges, while \( \mathit{sup} = 0 \) implies that missing edges are ignored and only deviations are penalized.
 In each recursion step, the algorithm selects the cut with the minimum cost, splits the event log accordingly, and recursively processes each resulting sub-log until only base cases remain.


\section{Domain-Enhanced Process Discovery with LLMs}\label{sec:framework}

The core idea of the proposed framework in this paper is to utilize domain knowledge to generate a set of rules $R$ which serves as input for the IMr framework. This can be done before starting the discovery by encoding process descriptions as rules, or after the process discovery by having a domain expert review the process model and provide feedback. Engaging in interactive conversations with LLMs in both scenarios helps address uncertainties and improve the quality of the extracted rules.

\begin{figure}[tb]
    \centering
    \includegraphics[width=0.9\linewidth]{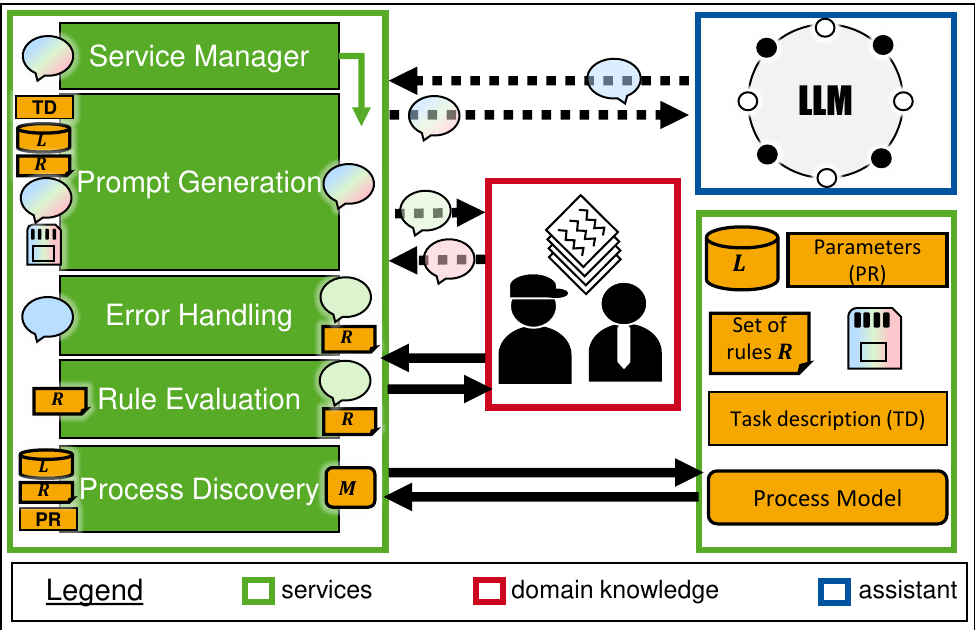}
    \caption{Overview of the proposed framework for LLM-assisted process discovery.}
    \label{fig:framework_HL}
\end{figure}

Figure~\ref{fig:framework_HL} presents an overview of the proposed framework, which models the interactions among three main components: the large language model, shown in the blue box; domain knowledge, represented in the red box; and the algorithmic components implemented on a computer system, shown in green. These algorithmic components are referred to as \emph{services} throughout the paper. Each service is depicted as a green box, with its inputs shown on the left and outputs on the right.

The orange elements represent input assets and system outputs that are stored internally and are thus enclosed within the green area. Dashed arrows indicate natural language messages exchanged either between services and the LLM, or between the services and domain experts. This section provides a detailed explanation of the framework’s components and their interactions.

\subsection{Message coordination and LLM task}

In our framework, multiple components can generate messages that are communicated to the LLM as part of the prompt. To distinguish the origin of each message, we define a set of roles corresponding to the entities responsible for message creation. Specifically, we define the set of roles as \( Role = \{\mathit{domain\mhyphen expert},\, \mathit{assistant},\, \mathit{service} \} \), where:

\begin{itemize}
    \item \( \mathit{domain\mhyphen expert} \) refers to the human user who possesses domain-specific knowledge and interacts with the system to guide or refine the process discovery task.
    
    \item \( \mathit{assistant} \) denotes the large language model, which engages in interactive reasoning, asks clarification questions, and extracts rules based on user input and contextual information.
    
    \item \( \mathit{service} \) refers to automated system components that operate independently of the domain expert. These components perform predefined functions, generate structured messages, or process inputs to produce outputs such as error notifications, statistical summaries, or process models.
\end{itemize}

This role classification enables clear attribution of each message within the interaction flow, thereby supporting transparency, accountability, and traceability throughout the process.

\begin{definition}[Message Structure]
Let \( \mathcal{W} \) be the set of all words and punctuation symbols in natural language, and let \( \mathcal{W}^* \) denote the set of all finite-length natural language texts. Let
{\small
\(
Role {=} \{ \mathit{domain\mhyphen expert},\, \mathit{assistant},\, \mathit{service} \}
\)}
be the set of roles. A message is defined as a pair \( m = (role, w) \in Role \times \mathcal{W}^* \), consisting of a role identifier and its associated text. We define \( \mathcal{M} = Role \times \mathcal{W}^* \) as the universe of all possible messages.
\end{definition}


Any message generated by the LLM, the domain expert, or one of the services is forwarded to the service manager that determines the appropriate downstream service to handle it. If a message needs to be sent to the LLM, the system constructs the corresponding prompt as described in Section~\ref{sec:services}, and forwards it to the LLM. Once a response is received, it is routed back to the service manager for further processing.

\begin{definition}[LLM Task Interface]
An \emph{LLM task} is a function
\[
\mathit{LLM} : \mathcal{M}^* \rightarrow \mathcal{W}^*
\]
that maps a sequence of role-tagged messages to a natural language response generated by the LLM.
\end{definition}

\subsection{Services}
\label{sec:services}
Upon receiving a message, the service manager inspects its content and makes one of the following decisions:
\begin{itemize}
    \item If the message is intended for the LLM, the service manager invokes the prompt generation service to enrich and prepare the input.
    
    \item If the message is a JSON-formatted response generated by the LLM, it is forwarded to the error checking service for validation.
    
    \item If the message contains extracted rules and corresponding evaluation metrics based on the event log, the user is prompted to review the rules and select those that are logically consistent and suitable for inclusion in the discovery process.
\end{itemize}

\subsubsection{Prompt Generation}
\label{sec:prompt_gen}
The prompt generation service is responsible for enriching the message with additional context, including the task description, the list of activities extracted from the event log, and the latest set of rules selected by the user (or an empty set if no rules have been generated or selected yet). The prompt generator also appends a history of previous interactions to provide contextual continuity. The output of the prompt generator is a sequence of messages 
\( Prompt = \langle M_0, M_L, M_R \rangle \cdot M_H \cdot \langle M \rangle \in \mathcal{M}^{*} \) composed of the following components:
\paragraph{Task Description} \( M_0 = (\mathit{service}, W_0) \) such that \( W_0 \in \mathcal{W}^* \) is a task description that defines the LLM's objectives.
To ensure reliable constraint extraction, the LLM is assigned the role of an assistant specifically responsible for identifying declarative constraints from input text. Its behavior is anchored through role-based prompting, which explicitly instructs the model to extract constraints, return them in strict JSON format, and request clarification in cases of ambiguity. A deterministic output specification is enforced by defining a rigid JSON structure and explicitly prohibiting explanations, markdown, or non-conforming formats, thereby reducing hallucinated outputs and ensuring compatibility with automated parsers. 

The prompt includes a specific set of supported constraint templates (e.g., $AtMost1$, $Precedence$, $CoExistence$) while forbidding unsupported constructs such as disjunctions and logical operators. 
Few-shot learning is implemented by the inclusion of labeled examples for each constraint type, as well as two extended examples consisting of multi-sentence textual descriptions with activity names and their corresponding expected sets of rules.
 These examples serve to operationalize the semantics of each constraint and support generalization across diverse phrasings. The prompt is designed to support multi-turn dialogue by instructing the LLM to ask clarification questions when input is incomplete or ambiguous, thereby improving robustness. 
Additionally, error prevention strategies are embedded through repeated formatting instructions and examples, which help minimize malformed JSON and other common output issues.


An abstracted version of the task description message is as follows:

\begin{tcolorbox}[enhanced,breakable,
  colback=green!5!white, 
  colframe=green!60!black,
  left=0.5mm, right=0.5mm]
\small
\textbf{Task Overview}\\
You are an \emph{assistant} tasked with translating free-text process descriptions into declarative process constraints.

\textbf{Your objectives are:}
\begin{enumerate}
    \item Identify declarative constraints expressed in the input text.
    \item Return the extracted constraints in strict JSON format (see specification below).
    \item If the input is ambiguous, request clarification before producing constraints.
\end{enumerate}

\textbf{JSON Output Format:}
\begin{tcolorbox}[colback=white!95!gray, colframe=gray!50!black, sharp corners, boxrule=0.3pt]
\centering
\textit{This section defines the fixed JSON schema used for constraint output.}
\end{tcolorbox}

\textbf{Supported Templates}
\begin{tcolorbox}[colback=white!95!gray, colframe=gray!50!black, sharp corners, boxrule=0.3pt]
\centering
\textit{This section lists the supported declarative templates along with their syntax and semantics.}
\end{tcolorbox}

\textbf{Common Mistakes and Guidelines}
\begin{tcolorbox}[colback=white!95!gray, colframe=gray!50!black, sharp corners, boxrule=0.3pt]
\centering
\textit{This section provides known error patterns and instructions to avoid incorrect outputs.}
\end{tcolorbox}

\textbf{Interaction Protocol}
\begin{enumerate}
    \item Upon receiving input, check for ambiguity in the process description.
    \item If ambiguity is found, ask clarification questions before constraint extraction.
    \item You may internally reason about the input, but do not include any reasoning or explanation in the output.
    \item Once confident, output only the final JSON object—no additional text or commentary.
\end{enumerate}

\textbf{Examples}
\begin{tcolorbox}[colback=white!95!gray, colframe=gray!50!black, sharp corners, boxrule=0.3pt]
\centering
\textit{This section contains fixed examples of input texts and their expected JSON outputs.}
\end{tcolorbox}
\end{tcolorbox}

\paragraph{Activities Report} \( M_L = (\mathit{service}, W_L) \) such that \( W_L  \in \mathcal{W}^* \) includes the list of activities extracted from the event log.
    \begin{tcolorbox}[enhanced,breakable,
  colback=green!5!white, 
  colframe=green!60!black,
  left=0.5mm, right=0.5mm]
\small
The list of activities used in this process are the following. Please only use these activities to generate constraints:

\begin{tcolorbox}[colback=white!95!gray, colframe=gray!50!black, sharp corners, boxrule=0.3pt]
\centering
\textit{The list of activities present in the event log is extracted by a dedicated service and included in this section.
}
\end{tcolorbox}
\end{tcolorbox}
\paragraph{Rules Report} \( M_R = (\mathit{service}, W_R) \) such that \( W_R \in \mathcal{W}^* \) includes the most recent list of rules selected by the domain expert. If no rules have been generated or selected, this message is omitted. 
\begin{tcolorbox}[enhanced,breakable,
  colback=green!5!white, 
  colframe=green!60!black,
  left=0.5mm, right=0.5mm]
\small
So far, the user has selected the following rules:
\begin{tcolorbox}[colback=white!95!gray, colframe=gray!50!black, sharp corners, boxrule=0.3pt]
\centering
\textit{The list of rules selected in the user interface is derived by a service and included here.
}
\end{tcolorbox}
\end{tcolorbox}
\paragraph{Historical Messages} \( M_H \in \mathcal{M}^* \) is the history of previous messages in the conversation. The history of the messages is always stores as an asset in the system as represnted with the memory disc icon in Fig.~\ref{fig:framework_HL}.
\paragraph{New Message} \( M = (\mathit{domain\mhyphen expert}, W) \) if \( W \in \mathcal{W}^* \) is the current input from the domain expert and \( M = (\mathit{service}, W) \) if \( W \in \mathcal{W}^* \) is the error explanation from the error checking service.

Each time a prompt is passed to the LLM, it returns a response \( A = \mathit{LLM}(Prompt) \in \mathcal{W}^* \). After each LLM invocation, the conversation history is updated as \( M_H = M_H \cdot \langle M \rangle \cdot \langle (\mathit{assistant}, A) \rangle \).


\subsubsection{Rule Validation}  
The declarative constraints are extracted from the LLM's response, which is expected to be a well-formed JSON object. The LLM is explicitly instructed to return only the constraints in a predefined JSON format, using a controlled vocabulary and without any additional text or explanation.
Following extraction, the constraints are subjected to a validation process to ensure their syntactic and semantic correctness. This includes verifying that each constraint matches the expected template structure, such as checking that the template identifier is valid and that the correct number of activity parameters is provided. Furthermore, each referenced activity label is checked against the set of activities present in the event log to ensure consistency.

If any issues are detected during validation, an error-handling loop is triggered. A diagnostic message is generated that describes the nature and location of the error, and a new prompt is constructed to inform the LLM of the issue. This allows the LLM to revise its output and produce a corrected version of the constraint set.

\subsubsection{Other Services}
If the LLM produces a set of rules in valid JSON format and the error validation returns no issues, the rule evaluation service computes the support and confidence of each extracted rule. The resulting statistics are then presented to the domain expert, who performs a final review and selects the rules that align with domain knowledge. This additional validation step provides an extra layer of safety, allowing the user to identify any illogical or implausible rules and, if necessary, revise their own explanations or resolve ambiguities, particularly when the statistical evidence contradicts their initial assumptions.

After the rules are extracted and the desired ones are selected, the domain expert can invoke the process discovery service, as described in Section~\ref{sec:IMr_intro}. The service generates a process model, which the domain expert can then review and validate. Based on this evaluation, the expert may provide further feedback or instructions to refine the model if necessary.

\subsection{Domain Expert}
\label{subsec:domain_expert}

The involvement of the domain expert in our framework can occur at two key stages: (1) prior to the discovery of a process model, where domain knowledge is introduced proactively, and (2) after an initial model is discovered, where expert feedback is used to refine and validate the model. Each interaction scenario supports different use cases and contributes to the overall quality of the discovered constraints.

\paragraph{Business Context Before Discovery}
In this setting, the domain expert introduces the actual business process to the LLM before any model is generated. This input may include a general description of the process, relationships between specific activities, or even declarative constraints formulated in natural language. Such knowledge may originate from formal process documentation, institutional know-how, or operational experience. By providing this context upfront, the domain expert can guide the discovery process and reduce the likelihood of irrelevant or misleading model structures.

\paragraph{Feedback Integration After Discovery}
A more common use case involves engaging the domain expert after the initial process model has been discovered. In this scenario, the model is presented to the expert for review, initiating an interactive dialogue aimed at improving its quality. The domain expert is expected to assess the model in terms of accuracy, completeness, and alignment with real-world process behavior. If discrepancies or implausible behaviors are identified, the expert provides written feedback describing the issues and clarifying the intended process semantics. Based on this feedback, the LLM is prompted to adjust the set of declarative constraints to better reflect the actual process.

\paragraph{Interactive Clarification and Refinement}
To support deeper understanding and refinement, the framework enables the LLM to express uncertainty and actively seek clarification. When ambiguities or gaps are detected in the input text, the LLM is encouraged to formulate follow-up questions, which are presented to the domain expert. The responses to these questions play a critical role in improving the precision of the extracted constraints. This interactive loop fosters a collaborative refinement process in which the LLM and domain expert jointly contribute to building a high-quality, context-aware process model.

After the successful extraction of rules, a service computes the \emph{support} and \emph{confidence} values for each rule with respect to the given event log. These metrics are then presented to the domain expert, who can decide whether to include or exclude individual rules based on this feedback. The numerical values serve as an important indicator of how well the extracted rules align with the actual event data, helping to identify potential compliance issues or inconsistencies between textual descriptions and observed behavior.

Once the expert selects the desired subset of rules through the interactive interface, the selected rules are passed to the IMr process discovery framework. There, they are used to guide the discovery of a process model that is not only data-driven but also informed by domain knowledge and aligned with expert expectations.

\section{Tool Support}
\label{sec:tool}
\begin{figure}[htb]
    \centering
    \includegraphics[width=0.8\linewidth]{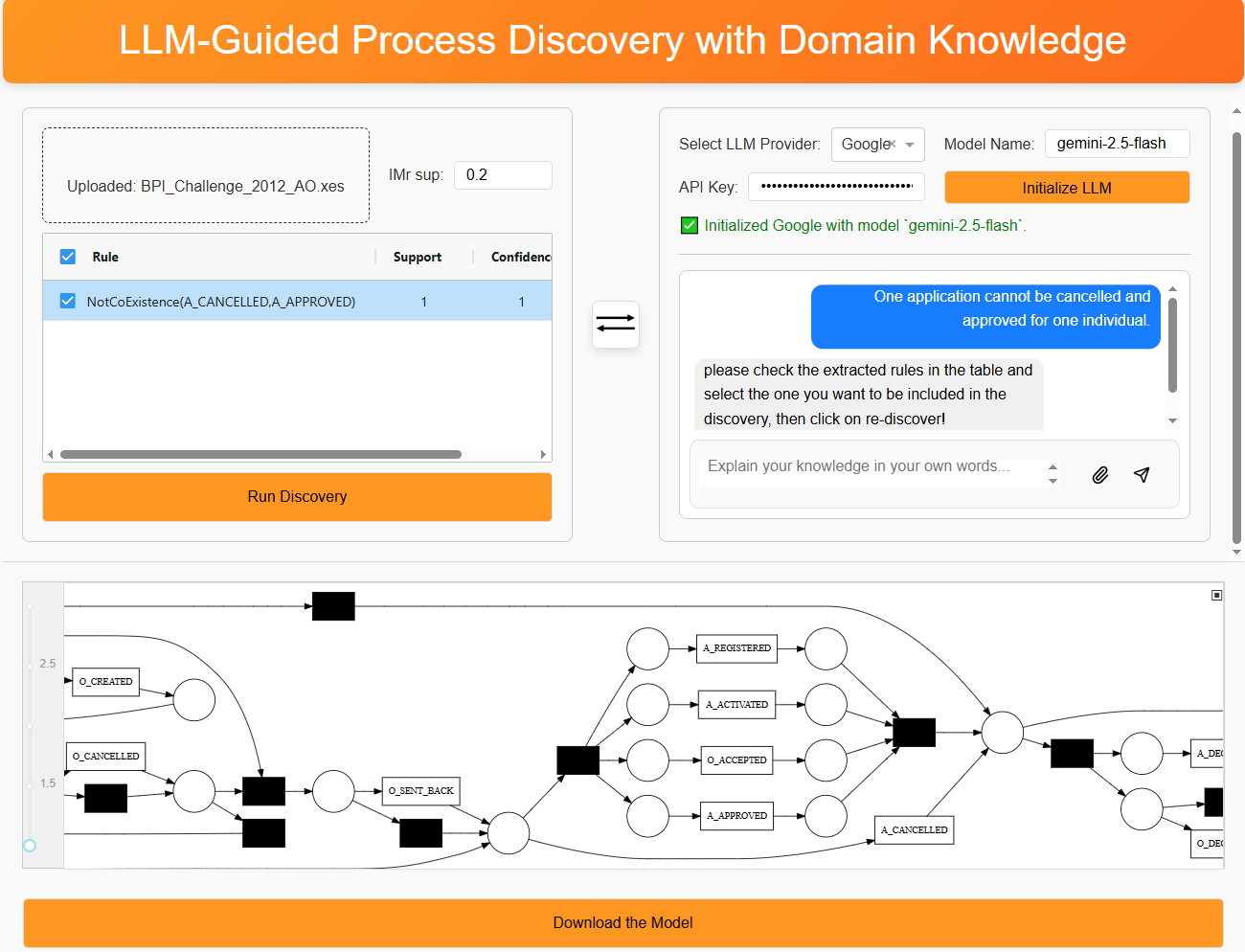}
    \caption{Screenshot of the interactive tool supporting the proposed framework.}
    \label{fig:tool_shot}
\end{figure}

An implementation of the proposed framework is publicly available\footnote{\url{https://github.com/aliNorouzifar/domain-guided-process-discovery}}. To enhance usability, we provide a web-based user interface that can be accessed in two ways: either by cloning the repository and installing the required packages manually, or more conveniently, by pulling and running a pre-built Docker image as described in the repository. The latter approach eliminates the need for manual dependency management and environment setup. Figure~\ref{fig:tool_shot} shows a screenshot of the developed tool, which operationalizes the framework and supports interactive collaboration between the domain expert, the LLM, and the internal services. All services depicted in Figure~\ref{fig:framework_HL} are implemented as backend components. Among them, the only service directly visible and configurable by the domain expert is the process discovery service.

The \emph{top section} of the user interface is divided into two main panels:

\begin{itemize}
    \item \textbf{Left Panel:} This panel allows the domain expert to upload an event log, configure the \emph{sup} parameter of the IMr discovery technique, and inspect the rules extracted by the LLM and validated through the error checking service. Notably, the user can initiate the discovery process without interacting with the LLM, in which case the process model is generated based on an empty set of rules. If rules are extracted from the LLM output, the backend services compute the \emph{support} and \emph{confidence} values for each rule based on the uploaded event log, using the Janus rule-checking tool (invoked as a service)\footnote{\url{https://github.com/Oneiroe/Janus}}. The domain expert can then select or deselect individual rules via checkboxes. Once the desired subset of rules is chosen, pressing the \emph{Run Discovery} button triggers the IMr process discovery algorithm. The resulting process model is displayed in the \emph{bottom section} of the interface. Any rule selected at this step will be appended to subsequent messages in the ongoing interaction.

    \item \textbf{Right Panel:} This panel is dedicated to model selection and LLM initialization. The user can choose from multiple LLM providers (e.g., OpenAI, Google, and DeepSeek), select a specific model from the chosen provider, and enter a valid API key. Upon clicking the \emph{Initialize LLM} button, a chat interface is activated, enabling live, multi-turn interaction with the selected LLM. This interface is supported by backend services responsible for prompt construction, validation, and error handling, as described in Section~\ref{sec:framework}. The dialogue continues until the LLM outputs a JSON-formatted file containing extracted rules. This output is then passed to the error validation service to ensure syntactic correctness. If any errors are detected, the backend services autonomously interact with the LLM to resolve the issues and request a corrected response. Once a valid set of rules is prepared, their quality is assessed using evaluation metrics such as support and confidence. These statistics are presented to the user, who is invited to review and select the rules that align with their domain knowledge. After finalizing the selection, the user can trigger the process discovery step by clicking the \emph{Run Discovery} button, resulting in the generation of a new process model.

\end{itemize}

The overall design of the tool is intended to streamline the integration of human knowledge and automated reasoning, making the process of rule extraction, validation, and guided process discovery both intuitive and traceable.

\section{Evaluation}
\label{sec:evaluation}
Our evaluation focuses on answering several key questions. First, we assess how well large language models can extract declarative rules from natural language input. Another objective of our evaluation is to test the interactive behavior of the framework. Specifically, we analyze whether the LLM correctly engages in clarification when the provided domain input is ambiguous or underspecified. For this, we intentionally introduce vague textual descriptions and examine the LLM’s responses.
Finally, we conduct a case study evaluation using real-life event logs from UWV, the Dutch employee insurance agency. In this case, we had the opportunity to collaborate directly with domain experts to elicit and incorporate domain knowledge into the discovery process.

\subsection{Prompting Strategies and LLM Comparison}
The proposed integration of large language models relies on receiving textual input from the user, with the LLM tasked with extracting declarative process constraints from this input. The user-provided text is not restricted to a single sentence, it may span multiple sentences or an entire paragraph. 
This part of the evaluation investigates how different prompting strategies and LLM configurations influence the extraction of constraints. Our objectives are as follows:

\begin{itemize}
    \item \textbf{Quantitative Evaluation:} We introduce a set of evaluation metrics to quantify the quality of LLM responses, enabling a systematic comparison of results across different prompting strategies and model configurations.

    \item \textbf{Model Performance Comparison:} We compare several state-of-the-art LLMs to assess whether their performance differs significantly in extracting declarative constraints from natural language inputs.
    
    \item \textbf{Effect of Few-Shot Prompting:} We evaluate whether augmenting the task description with a few examples (few-shot prompting) improves the quality of extracted constraints compared to using only the task description (zero-shot prompting).
    
    \item \textbf{Input Granularity Analysis:} We examine whether there is a notable difference in LLM performance when prompts contain one sentence per desired constraint versus a full paragraph describing multiple constraints.

\end{itemize}


\subsubsection{Experimental Setup}
We selected several real-life event logs for which a body of published research exists and a shared understanding of the underlying processes has emerged within the process mining community~\cite{DBLP:conf/bpm/LopesF19}. Specifically, we used the BPI Challenge 2017 log (BPIC17), which pertains to a loan application process, the BPI Challenge 2018 log (BPIC18), which relates to the allocation of agricultural subsidies to European farmers, and the UWV event log extracted from a claim handling process in an insurance agency, for which we had direct access to domain experts at the agency.

\begin{figure}[htb]
\centering 
\begin{subfigure}{1\textwidth}
    \centerline{\includegraphics[scale=0.1]{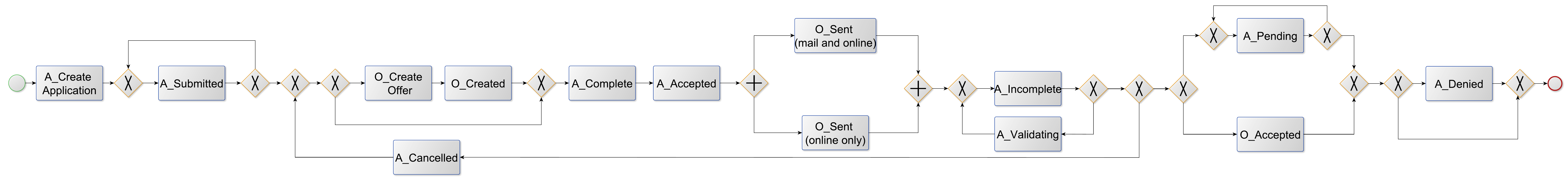}}
    \caption{Manually constructed model for the BPIC17 event log, representing a loan application process with intentional violations of constraints \(d_1\) to \(d_{10}\).}
    \label{fig:BPIC17_test}
\end{subfigure}\\
\begin{subfigure}{1\textwidth}
    \centerline{\includegraphics[scale=0.1]{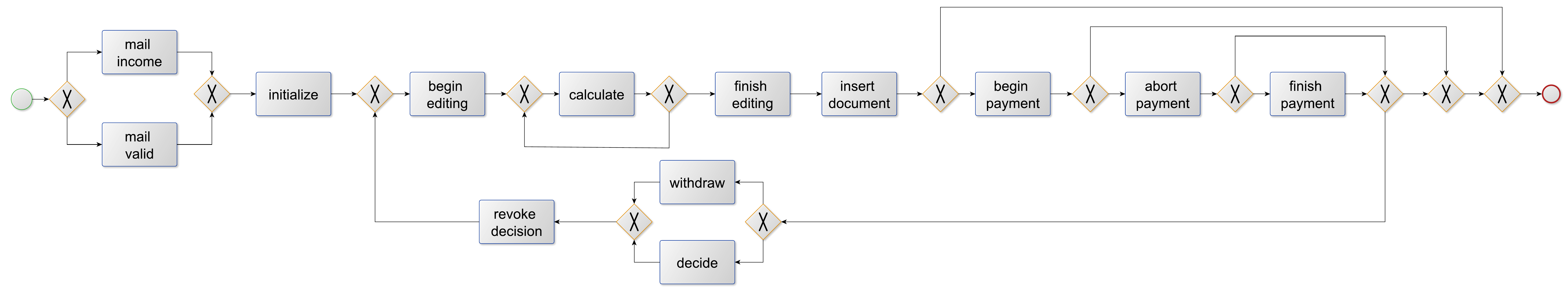}}
    \caption{Manually constructed model for the BPIC18 event log, corresponding to an agricultural subsidy process, with embedded deviations from constraints \(d_{11}\) to \(d_{20}\).}
    \label{fig:BPIC18_test}
\end{subfigure}\\
\begin{subfigure}{1\textwidth}
    \centerline{\includegraphics[scale=0.1]{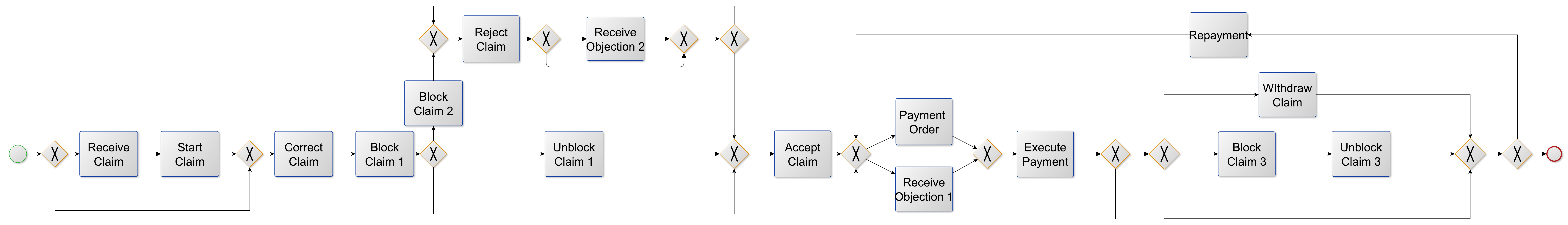}}
    \caption{Manually constructed model for the UWV event log from a claim handling process, intentionally violating constraints \(d_{21}\) to \(d_{31}\).}
    \label{fig:UWV_test}
\end{subfigure}
\caption{Manually constructed process models used to evaluate the ability of LLMs to detect violations based on natural language descriptions of constraints. Each model contains known deviations from a predefined set of declarative rules.}
\end{figure}

\paragraph{Data}
For each event log, we manually constructed process models that are deliberately designed to include behavioral deviations. These models are shown in Fig.~\ref{fig:BPIC17_test}, Fig.~\ref{fig:BPIC18_test}, and Fig.~\ref{fig:UWV_test}, corresponding to the BPIC17, BPIC18, and UWV event logs, respectively. For each model, a predefined set of declarative rules, violated by the respective model, is generated. A selection of these rules is presented in Table~\ref{tab:sentences_examples}, where rules \( d_1 \) to \( d_{10} \) correspond to BPIC17, \( d_{11} \) to \( d_{20} \) to BPIC18, and \( d_{21} \) to \( d_{31} \) to the UWV event log. These rules are assumed to hold in compliant process executions. The objective of this evaluation is to assess whether the LLM can correctly identify these deviations when provided with natural language descriptions of the expected behavior.

For each constraint \( d_i \),
we crafted three semantically equivalent sentences \( s_{i,1} \), \( s_{i,2} \), and \( s_{i,3} \) each using different phrasings. This variation is intended to capture the diversity in how domain experts may naturally express the same constraint. Table~\ref{tab:sentences_examples} presents a selection of declarative constraints along with representative sentences formulated to express them in natural language. 

\begin{table}[H]
\footnotesize
\centering
\caption{Examples of natural language formulations for a selection of declarative constraints (\(d_1\) to \(d_{31}\)) drawn from the BPIC17, BPIC18, and UWV event logs. Each constraint is expressed through multiple representative sentences, reflecting how domain experts might describe the intended behavior of a process. }
\label{tab:sentences_examples}
\begin{tabular}{|p{4cm}|p{4cm}|p{4cm}|}
\hline
\multicolumn{3}{|c|}{\textbf{Declarative Rule}} \\ \hline

\textbf{Sentence 1} & \textbf{Sentence 2} & \textbf{Sentence 3} \\ \hline

\multicolumn{3}{|c|}{\( d_1 \): \emph{AtLeast1(O\_Create Offer)} } \\ \hline
In all applications, the bank is required to create at least one offer. & Creating an offer is a mandatory step in the loan application process. & It is not possible to have instances without any offers created. \\ \hline

... & ... & ... \\ \hline

\multicolumn{3}{|c|}{\( d_{10} \): \emph{Precedence(O\_Accepted, A\_Pending)} } \\ \hline
If an application is in pending, there must be an accepted offer beforehand. & All pending applications must be preceded by an accepted offer. & It should not be possible to have a pending application without first having an offer accepted. \\ \hline \hline

\multicolumn{3}{|c|}{\( d_{11} \): \emph{NotSuccession(decide, begin editing)} } \\ \hline
After making a decision, no editing should begin. & Begin edition should not be permitted after making a decision. & We cannot begin editing when a decision is made. \\ \hline

... & ... & ... \\ \hline

\multicolumn{3}{|c|}{\( d_{21} \): \emph{AtMost1(withdraw)} } \\ \hline
Withdrawing the case cannot occur more than once. & Each application can be withdrawn a maximum of one time. & We cannot have multiple withdrawals within the same application. \\ \hline \hline

\multicolumn{3}{|c|}{\( d_{22} \): \emph{NotCoExistence(Accept Claim, Reject Claim)} } \\ \hline
One claim can be either accepted or rejected. & We cannot have claims that are both accepted and rejected. & If a claim is accepted, it cannot be rejected and vice versa. \\ \hline

... & ... & ... \\ \hline

\multicolumn{3}{|c|}{\( d_{31} \): \emph{Response(Unblock Claim 3, Payment Order)} } \\ \hline
It is not possible to have the third unblocking mechanism event without any payment order occurring after it. & Unblocking claims with the third mechanism must be followed by a payment order. & If a claim is unblocked with type three, then a payment order must occur afterward. \\ \hline
\end{tabular}
\end{table}

\paragraph{Configurations}
Given a prompt structure defined as \( \mathit{Prompt} = \langle M_0, M_L, M_R \rangle \cdot M_H \cdot \langle M \rangle \in \mathcal{M}^{*} \), as introduced in Section~\ref{sec:prompt_gen}, we apply specific modifications to enable controlled comparisons between different prompting strategies and LLM configurations. In this evaluation setting, we assume no rules have been preselected, and there is no prior interaction history. Consequently, the message components \( M_R \) and \( M_H \) are empty. $M_L$ is automatically generated for each individual event log and includes a list of activities integrated in a message.

Three different configurations are tested in our evaluation including:
\begin{itemize}
    \item \textbf{Different LLM models}: To evaluate the effectiveness of different language models in extracting declarative rules from textual inputs, we tested several state-of-the-art LLMs from three major providers. Specifically, we included \textbf{Gemini 2.5 Flash} and \textbf{Gemini 2.5 Pro} from Google, \textbf{GPT-4.1 Nano}, \textbf{GPT-4.1}, and \textbf{GPT-o3} from OpenAI, as well as \textbf{DeepSeek Chat} (DeepSeek-V3-0324) and \textbf{DeepSeek Reasoner} (DeepSeek-R1-0528) from DeepSeek. These models are evaluated and compared across identical prompting conditions to assess their performance in identifying declarative constraints from natural language descriptions.
    \item \textbf{Few-shot vs zero-shot prompting}: We introduce two variations of the task description prompt \( M_0 \). The first variant, denoted \( M_0' \), retains the original task description but omits the instruction to ask clarification questions. In this configuration, the LLM is explicitly instructed to extract declarative rules from the input. This setup corresponds to a \emph{few-shot prompting} strategy, as it includes example pairs of inputs and expected outputs. The second variant, denoted \( M_0'' \), further simplifies the prompt by removing both the questioning instruction and the example pairs. Here, only the task description is provided, corresponding to a \emph{zero-shot prompting} setup. These two configurations allow us to isolate the impact of the prompting strategy on rule extraction performance across different LLMs.
    \item \textbf{Sentence to sentence vs paragraph level prompting}: Let \( I \subseteq \mathcal{P}(W^*) \) denote the set of input text blocks used for evaluating the LLM's rule extraction capabilities. We consider two prompting configurations. In the \emph{sentence-to-sentence} (S2S) setting, each prompt contains exactly one sentence, and the LLM is expected to extract a single declarative rule per input. This setup isolates each constraint description, allowing for fine-grained evaluation. The input set in this case is defined as
\(
I = \{ s_{i,j} \mid 1 \leq i \leq 31,\ 1 \leq j \leq 3 \}
\). The \emph{paragraph-level} (PAR) configuration aggregates multiple sentences, each describing a different constraint, into a single prompt, allowing the LLM to extract a set of rules jointly. In this case, the input set is
\(
I = \{ p_{\ell,j} \mid \ell \in \{\text{BPIC17}, \text{BPIC18}, \text{UWV}\},\ 1 \leq j \leq 3 \}
\)
where each \( p_{\ell,j} \) is a paragraph formed by concatenating multiple rule-descriptive sentences for dataset \( \ell \). Specifically:
\(
p_{\text{BPIC17},j} = s_{1,j} \cdot \ldots \cdot s_{10,j},
p_{\text{BPIC18},j} = s_{11,j} \cdot \ldots \cdot s_{20,j},
p_{\text{UWV},j}    = s_{21,j} \cdot \ldots \cdot s_{31,j}
\).

\end{itemize}

\paragraph{Evaluation Metrics}
 The proposed LLM-based method then attempts to extract rules from the input set of text blocks $I$, and its performance is assessed by comparing the results against the annotated ground truth using standard information retrieval metrics. For each $w \in I$, we consider a ground truth set of rules and the set of rules extracted by LLM as: 

\begin{itemize}
    \item \( \mathit{ground\mhyphen truth}(w)\) returns the set of declarative rules manually identified as correct for a given text block. 
    
    \item \( \mathit{LLM\mhyphen rules}(w)\) returns the set of declarative rules extracted by the LLM for a given text block.
\end{itemize}

 We define the following evaluation metrics:
\begin{itemize}
    \item \textbf{Recall} measures the proportion of ground truth rules that are correctly extracted by the LLM. It penalizes omissions:
    \[
    \mathit{recall} = \frac{\sum_{w \in I} \left| \mathit{ground\mhyphen truth}(w) \cap \mathit{LLM\mhyphen rules}(w) \right|}{\sum_{w \in I} \left| \mathit{ground\mhyphen truth}(w) \right|}
    \]

    \item \textbf{Precision} measures the proportion of extracted rules that match the ground truth. It penalizes hallucinated or incorrect rules:
    \[
    \mathit{precision} = \frac{\sum_{w \in I}\left| \mathit{ground\mhyphen truth}(w) \cap \mathit{LLM\mhyphen rules}(w) \right|}{\sum_{w \in I} \left| \mathit{LLM\mhyphen rules}(w) \right|}
    \]

    \item \textbf{Error Rate} is defined as the fraction of prompts that triggered at least one error, thereby requiring activation of the error-handling mechanism.

    \item \textbf{Failure Rate} denotes the fraction of prompts for which the LLM failed to produce any valid rule after a maximum of 10 error-correction iterations, resulting in termination without output.
\end{itemize}

\subsubsection{Experimental Results}
The generated data were evaluated under various configurations, and the corresponding performance metrics are reported in Table~\ref{tab:recall}. Considering different LLMs, the highest recall and precision values across all models are highlighted in bold. Additionally, for each specific LLM, the configuration yielding its highest recall and precision is marked in green to emphasize the best-performing setup per model.

\begin{table}[htb]
\centering
\caption{Quantitative evaluation of different LLM models and configurations on rule extraction from natural language inputs.}
\label{tab:recall}
\begin{tabular}{ccl|l|l|l|l|l|l|l|}
\cline{4-10}
\multicolumn{1}{l}{}                             & \multicolumn{1}{l}{}                       &              & \rotatebox{90}{Gemini 2.5 flash} & \rotatebox{90}{Gemini 2.5 pro} & \rotatebox{90}{OpenAI 4.1 nano} & \rotatebox{90}{OpenAI 4.1} & \rotatebox{90}{OpenAI o3} & \rotatebox{90}{Deepseek chat} & \rotatebox{90}{Deepseek reasoner} \\ \hline
\multicolumn{1}{|c|}{}                            & \multicolumn{1}{c|}{}                      & recall       & 0.92                         & \cellcolor[HTML]{67FD9A}\textbf{0.98} & 0.53                         & 0.90                                  & 0.96                                  & 0.85                         & 0.94                         \\ \cline{3-10} 
\multicolumn{1}{|c|}{}                            & \multicolumn{1}{c|}{}                      & precision    & 0.76                         & \textbf{0.77}                         & 0.45                         & \cellcolor[HTML]{67FD9A}\textbf{0.77} & 0.73                                  & 0.68                         & 0.69                         \\ \cline{3-10} 
\multicolumn{1}{|c|}{}                            & \multicolumn{1}{c|}{}                      & error rate   & 0.01                         & 0                                     & 0.40                         & 0.01                                  & 0                                     & 0                            & 0                            \\ \cline{3-10} 
\multicolumn{1}{|c|}{}                            & \multicolumn{1}{c|}{\multirow{-4}{*}{S2S}} & failure rate & 0                            & 0                                     & 0.01                         & 0                                     & 0                                     & 0                            & 0                            \\ \cline{2-10} 
\multicolumn{1}{|c|}{}                            & \multicolumn{1}{c|}{}                      & recall       & 0.91                         & {\color[HTML]{000000} \textbf{0.95}}  & 0.31                         & 0.59                                  & \textbf{0.95}                         & 0.83                         & 0.94                         \\ \cline{3-10} 
\multicolumn{1}{|c|}{}                            & \multicolumn{1}{c|}{}                      & precision    & 0.75                         & {\color[HTML]{000000} \textbf{0.87}}  & 0.34                         & 0.83                                  & 0.83                                  & 0.81                         & 0.76                         \\ \cline{3-10} 
\multicolumn{1}{|c|}{}                            & \multicolumn{1}{c|}{}                      & error rate   & 0                            & 0.02                                  & 0.04                         & 0.04                                  & 0                                     & 0                            & 0                            \\ \cline{3-10} 
\multicolumn{1}{|c|}{\multirow{-8}{*}{zero shot}} & \multicolumn{1}{c|}{\multirow{-4}{*}{PAR}} & failure rate & 0                            & 0                                     & 0.01                         & 0.03                                  & 0                                     & 0                            & 0                            \\ \hline
\multicolumn{1}{|c|}{}                            & \multicolumn{1}{c|}{}                      & recall       & \cellcolor[HTML]{67FD9A}0.95 & \cellcolor[HTML]{67FD9A}\textbf{0.98} & \cellcolor[HTML]{67FD9A}0.62 & \cellcolor[HTML]{67FD9A}0.92          & \cellcolor[HTML]{67FD9A}\textbf{0.98} & \cellcolor[HTML]{67FD9A}0.92 & \cellcolor[HTML]{67FD9A}0.97 \\ \cline{3-10} 
\multicolumn{1}{|c|}{}                            & \multicolumn{1}{c|}{}                      & precision    & \textbf{0.81}                & 0.76                                  & \cellcolor[HTML]{67FD9A}0.59 & \cellcolor[HTML]{67FD9A}0.77          & 0.75                                  & 0.74                         & 0.78                         \\ \cline{3-10} 
\multicolumn{1}{|c|}{}                            & \multicolumn{1}{c|}{}                      & error rate   & 0                            & 0                                     & 0.22                         & 0                                     & 0                                     & 0                            & 0                            \\ \cline{3-10} 
\multicolumn{1}{|c|}{}                            & \multicolumn{1}{c|}{\multirow{-4}{*}{S2S}} & failure rate & 0                            & 0                                     & 0                            & 0                                     & 0                                     & 0                            & 0                            \\ \cline{2-10} 
\multicolumn{1}{|c|}{}                            & \multicolumn{1}{c|}{}                      & recall       & \cellcolor[HTML]{67FD9A}0.95 & 0.95                                  & 0.38                         & 0.82                                  & \textbf{0.97}                         & 0.85                         & 0.96                         \\ \cline{3-10} 
\multicolumn{1}{|c|}{}                            & \multicolumn{1}{c|}{}                      & precision    & \cellcolor[HTML]{67FD9A}0.86 & \cellcolor[HTML]{67FD9A}\textbf{0.89} & 0.36                         & 0.72                                  & \cellcolor[HTML]{67FD9A}0.83          & \cellcolor[HTML]{67FD9A}0.83 & \cellcolor[HTML]{67FD9A}0.87 \\ \cline{3-10} 
\multicolumn{1}{|c|}{}                            & \multicolumn{1}{c|}{}                      & error rate   & 0.11                         & 0                                     & 0.44                         & 0                                     & 0                                     & 0                            & 0                            \\ \cline{3-10} 
\multicolumn{1}{|c|}{\multirow{-8}{*}{few shot}}  & \multicolumn{1}{c|}{\multirow{-4}{*}{PAR}} & failure rate & 0                            & 0                                     & 0                            & 0                                     & 0                                     & 0                            & 0                            \\ \hline
\end{tabular}
\end{table}

\paragraph{Comparison Between LLM Models:} Among the evaluated models, \textbf{OpenAI 4.1 Nano} consistently exhibited the weakest performance. As a lightweight and cost-efficient variant of GPT-4, 4.1 Nano is optimized for speed and resource constraints rather than reasoning depth. This limitation is reflected in its results, with the lowest \emph{recall} and \emph{precision} across all configurations. Moreover, it showed a significantly higher \emph{error rate} and \emph{failure rate}, indicating limited robustness in syntactic and semantic rule extraction.

In contrast, \textbf{Gemini 2.5 Pro} and \textbf{OpenAI o3} demonstrated superior capabilities. These models, categorized as advanced reasoning variants, are designed to handle multi-step logic and complex semantic tasks. This is evident in their high \emph{recall} values, suggesting a strong ability to correctly identify and reproduce declarative constraints aligned with the ground truth. Their \emph{precision} scores are also relatively high, indicating that most of the extracted rules were correct and not hallucinated.
While other models such as \textbf{DeepSeek Reasoner}, \textbf{Gemini 2.5 Flash}, and \textbf{OpenAI 4.1} occasionally outperformed in specific configurations, especially in terms of precision, these gains are not consistent across all metrics.

Overall, the results support a clear preference for \textbf{Gemini 2.5 Pro} and \textbf{OpenAI o3} in scenarios that demand accurate and comprehensive extraction of declarative rules from natural language. Their balance of recall, precision, and reliability makes them particularly suitable for integration into our rule-guided process discovery pipeline.

\paragraph{Comparison of zero-Shot and few-Shot Prompting:} 
The results presented in Table~\ref{tab:recall} reveal a clear benefit from adopting few-shot prompting strategies. On average, we observe an increase of approximately \textbf{5\% in recall} and \textbf{4\% in precision} when moving from zero-shot to few-shot prompting across all experiments. This gain underscores the value of providing example-based guidance alongside task descriptions, especially for models that lack strong internal reasoning mechanisms.

When we isolate the more advanced reasoner models, namely, \textbf{OpenAI o3}, \textbf{Gemini 2.5 Pro}, and \textbf{DeepSeek Reasoner}, the improvement in recall is relatively modest, averaging around \textbf{1\%}. In contrast, other models exhibit a more substantial gain of approximately \textbf{7\%} in recall. This contrast highlights that few-shot prompting particularly benefits models that do not inherently excel at multi-step logical reasoning or pattern generalization.

The precision gains are more uniformly distributed across both groups, with an average improvement of \textbf{around 4\%} in both advanced reasoner models and other models. This suggests that few-shot prompting not only helps to recall more ground truth rules but also enhances the reliability of the extracted rules by reducing hallucinations.


\paragraph{Comparison Between S2S and PAR Input:}
Although both the \emph{S2S} (sentence-to-sentence) and \emph{PAR} (paragraph-level) configurations use the same set of sentences and are evaluated against the same ground truth, the results reveal meaningful differences in performance. In the \emph{S2S} setup, where each sentence is processed individually in a separate prompt, we observe an average improvement of approximately \textbf{8\% in recall}. This suggests that presenting each sentence in isolation enables the LLMs to focus more precisely on the intended meaning and correctly extract the associated declarative constraint, leading to higher rule coverage.

Conversely, the \emph{PAR} configuration—where multiple sentences are concatenated and processed jointly in a single prompt—achieves \textbf{around 4\% higher precision} on average. This improvement likely stems from the richer context available to the LLM and the reduced number of prompt invocations, which together help limit the generation of spurious or overly general rules not present in the ground truth. These findings indicate a trade-off: \emph{S2S} favors completeness (higher recall), while \emph{PAR} favors precision. Depending on the application, whether maximizing coverage or minimizing false positives is more critical, one configuration may be more suitable than the other.

\subsection{Handling Ambiguities}
To evaluate the LLMs' ability to handle ambiguous input, we manually constructed a set of 12 sentences that intentionally contain vague or underspecified information. The objective of this experiment is to assess whether the LLMs appropriately enter a question-and-answer mode—seeking clarification from the user—or whether they attempt to generate declarative rules despite lacking essential domain knowledge. For each event log, we created 4 such ambiguous sentences, as detailed below:

\begin{itemize}
    \item \textbf{S1:} (BPIC17) \textit{Validating an application occurs at most two times.} \\
    \textbf{Reason:} The constraint template \emph{AtMost2}, which restricts the number of occurrences of an activity to two, is not included in the instruction prompt. As a result, the LLM lacks the necessary schema to map this statement to a valid declarative constraint.

    \item \textbf{S2:} (BPIC17) \textit{Each time an offer is created, it is sent via mail only.} \\
    \textbf{Reason:} The activity \texttt{O\_Sent (mail only)} does not exist in the BPIC17 event log. Instead, the log contains \texttt{O\_Sent (online only)} and \texttt{O\_Sent (mail and online)}. This discrepancy renders the statement ambiguous or factually incorrect with respect to the available activity set.

    \item \textbf{S3:} (BPIC17) \textit{The loan application process has three possible outcomes that cannot coexist for a specific application.} \\
    \textbf{Reason:} While the sentence refers to mutually exclusive outcomes, it does not explicitly mention the corresponding activity labels. Without these, the LLM cannot infer the specific constraints (e.g., \emph{NotCoExistence}) that need to be derived from the statement.

    \item \textbf{S4:} (BPIC17) \textit{In certain rare situations, it has been observed that the cancellation of an application may be followed by the creation of another offer.} \\
    \textbf{Reason:} The sentence uses hedging language (e.g., ``may", ``in rare situations") which introduces vagueness. Declarative process constraints require definitive assertions about behavioral regularities—statements must be clearly framed as either expected or forbidden. The fuzzy phrasing inhibits reliable constraint extraction.

    \item \textbf{S5:} (UWV) \textit{One of the blocking mechanisms of the claims only occurs for the accepted claims.} \\
    \textbf{Reason:} The UWV event log contains three blocking activities—\texttt{Block Claim 1}, \texttt{Block Claim 2}, and \texttt{Block Claim 3}. The statement does not specify which of these is intended, making it impossible to determine the correct activity for constraint formulation.

    \item \textbf{S6:} (UWV) \textit{A claim should not be subject to two conflicting actions.} \\
    \textbf{Reason:} The phrase “conflicting actions” is ambiguous. Without a precise definition or enumeration of the conflicting activities, the LLM cannot reliably map the statement to a specific constraint type such as \texttt{NotCoExistence} or \texttt{NotSuccession}.

    \item \textbf{S7:} (UWV) \textit{Receiving an objection can occur a maximum of once.} \\
    \textbf{Reason:} The event log distinguishes between two types of objections—\texttt{Receive Objection 1} and \texttt{Receive Objection 2}. The sentence does not clarify whether the limit applies to one specific activity or to both combined.

    \item \textbf{S8:} (UWV) \textit{In some cases, once an application is rejected, further actions may not be expected.} \\
    \textbf{Reason:} The use of hedging language (e.g., ``in some cases", ``may not be expected") makes the intent unclear. Declarative constraints require definitive behavioral assertions, and such probabilistic or conditional language cannot be directly encoded in the constraint formalism.

    \item \textbf{S9:} (BPIC18) \textit{Abort editing cannot occur after finishing the editing.} \\
    \textbf{Reason:} There is no activity labeled \texttt{Abort Editing} in the event log. However, the presence of a similarly named activity, \texttt{Abort Payment}, may lead to confusion or misinterpretation by the LLM.

    \item \textbf{S10:} (BPIC18) \textit{All payments may eventually be finished.} \\
    \textbf{Reason:} The statement uses weak and non-committal language (e.g., ``may eventually"), making it unclear whether this is intended as a strict behavioral requirement. Declarative constraints require clearly defined temporal or occurrence-based rules, which this sentence lacks—similar to S4 and S8.

    \item \textbf{S11:} (BPIC18) \textit{Calculations must occur after editing.} \\
    \textbf{Reason:} The event log includes two separate activities related to editing—\texttt{Begin Editing} and \texttt{Finish Editing}. The sentence does not specify which of the two should precede the \texttt{Calculation} activity, making the intended precedence relation ambiguous.

    \item \textbf{S12:} (BPIC18) \textit{Making a decision and revoking a decision can sometimes coexist.} \\
    \textbf{Reason:} The sentence uses vague quantifiers such as ``sometimes,” which do not translate into well-defined declarative semantics. As with S4, S8, and S10, declarative languages do not support probabilistic or fuzzy expressions of coexistence.
\end{itemize}


Different LLMs are evaluated on their ability to extract declarative rules from a set of deliberately ambiguous sentences. Each model is explicitly instructed to request clarification through follow-up questions whenever the input text exhibits ambiguity or lacks sufficient detail. Table~\ref{tab:vague} summarizes the results using a color-coded scheme: 

\begin{itemize}
    \item \textbf{Gray:} The model returned an empty set of rules without posing any follow-up questions.
    \item \textbf{Green:} The model asked a relevant and context-aware clarification question that directly addressed the ambiguity in the input sentence.
    \item \textbf{Yellow:} The model asked a question, but it was either irrelevant or unrelated to the actual source of ambiguity in the input.
    \item \textbf{Red:} The model produced a list of constraints without raising any questions, despite the presence of clear ambiguities in the input.
\end{itemize}

This evaluation highlights how well each model recognizes uncertainty in user input and whether it is capable of engaging in interactive clarification before extracting declarative constraints.

\begin{table}[htb]
\centering
\caption{Evaluation of LLM responses to deliberately ambiguous sentences.}
\label{tab:vague}
\begin{tabular}{l|l|l|l|l|l|l|l|l|l|l|l|l|}
\cline{2-13}
                                        & s1                                                       & s2                                & s3                                & s4                                & s5                                                       & s6                                                       & s7                                                       & s8                                                       & s9                                                       & s10                                                      & s11                                                      & s12                               \\ \hline
\multicolumn{1}{|l|}{Gemini 2.5 flash}  & \cellcolor[HTML]{C0C0C0}{\color[HTML]{000000} \textbf{}} & \cellcolor[HTML]{FE0000}\textbf{} & \cellcolor[HTML]{34FF34}\textbf{} & \cellcolor[HTML]{34FF34}\textbf{} & \cellcolor[HTML]{FE0000}\textbf{}                        & \cellcolor[HTML]{FE0000}\textbf{}                        & \cellcolor[HTML]{34FF34}{\color[HTML]{FE0000} \textbf{}} & \cellcolor[HTML]{F8FF00}{\color[HTML]{FE0000} \textbf{}} & \cellcolor[HTML]{34FF34}{\color[HTML]{FE0000} \textbf{}} & \cellcolor[HTML]{FE0000}\textbf{}                        & \cellcolor[HTML]{FE0000}\textbf{}                        & \cellcolor[HTML]{C0C0C0}\textbf{} \\ \hline
\multicolumn{1}{|l|}{Gemini 2.5 pro}    & \cellcolor[HTML]{34FF34}\textbf{}                        & \cellcolor[HTML]{FE0000}\textbf{} & \cellcolor[HTML]{34FF34}\textbf{} & \cellcolor[HTML]{C0C0C0}\textbf{} & \cellcolor[HTML]{34FF34}{\color[HTML]{FE0000} \textbf{}} & \cellcolor[HTML]{34FF34}{\color[HTML]{FE0000} \textbf{}} & \cellcolor[HTML]{FE0000}\textbf{}                        & \cellcolor[HTML]{FE0000}\textbf{}                        & \cellcolor[HTML]{FE0000}\textbf{}                        & \cellcolor[HTML]{FE0000}\textbf{}                        & \cellcolor[HTML]{FE0000}\textbf{}                        & \cellcolor[HTML]{C0C0C0}\textbf{} \\ \hline
\multicolumn{1}{|l|}{OpenAI 4.1 nano}   & \cellcolor[HTML]{FE0000}\textbf{}                        & \cellcolor[HTML]{FE0000}\textbf{} & \cellcolor[HTML]{FE0000}\textbf{} & \cellcolor[HTML]{F8FF00}\textbf{} & \cellcolor[HTML]{FE0000}\textbf{}                        & \cellcolor[HTML]{FE0000}\textbf{}                        & \cellcolor[HTML]{FE0000}\textbf{}                        & \cellcolor[HTML]{FE0000}\textbf{}                        & \cellcolor[HTML]{FE0000}\textbf{}                        & \cellcolor[HTML]{FE0000}\textbf{}                        & \cellcolor[HTML]{FE0000}\textbf{}                        & \cellcolor[HTML]{FE0000}\textbf{} \\ \hline
\multicolumn{1}{|l|}{OpenAI 4.1}        & \cellcolor[HTML]{C0C0C0}\textbf{}                        & \cellcolor[HTML]{FE0000}\textbf{} & \cellcolor[HTML]{FE0000}\textbf{} & \cellcolor[HTML]{C0C0C0}\textbf{} & \cellcolor[HTML]{FE0000}\textbf{}                        & \cellcolor[HTML]{FE0000}\textbf{}                        & \cellcolor[HTML]{FE0000}\textbf{}                        & \cellcolor[HTML]{FE0000}\textbf{}                        & \cellcolor[HTML]{FE0000}\textbf{}                        & \cellcolor[HTML]{FE0000}\textbf{}                        & \cellcolor[HTML]{FE0000}\textbf{}                        & \cellcolor[HTML]{C0C0C0}\textbf{} \\ \hline
\multicolumn{1}{|l|}{OpenAI o3}         & \cellcolor[HTML]{34FF34}\textbf{}                        & \cellcolor[HTML]{FE0000}\textbf{} & \cellcolor[HTML]{34FF34}\textbf{} & \cellcolor[HTML]{C0C0C0}\textbf{} & \cellcolor[HTML]{34FF34}\textbf{}                        & \cellcolor[HTML]{34FF34}{\color[HTML]{FE0000} \textbf{}} & \cellcolor[HTML]{34FF34}{\color[HTML]{FE0000} \textbf{}} & \cellcolor[HTML]{34FF34}{\color[HTML]{FE0000} \textbf{}} & \cellcolor[HTML]{34FF34}{\color[HTML]{FE0000} \textbf{}} & \cellcolor[HTML]{34FF34}{\color[HTML]{FE0000} \textbf{}} & \cellcolor[HTML]{34FF34}{\color[HTML]{FE0000} \textbf{}} & \cellcolor[HTML]{C0C0C0}\textbf{} \\ \hline
\multicolumn{1}{|l|}{Deepseek chat}     & \cellcolor[HTML]{FE0000}\textbf{}                        & \cellcolor[HTML]{FE0000}\textbf{} & \cellcolor[HTML]{FE0000}\textbf{} & \cellcolor[HTML]{FE0000}\textbf{} & \cellcolor[HTML]{FE0000}\textbf{}                        & \cellcolor[HTML]{FE0000}\textbf{}                        & \cellcolor[HTML]{FE0000}\textbf{}                        & \cellcolor[HTML]{FE0000}\textbf{}                        & \cellcolor[HTML]{FE0000}\textbf{}                        & \cellcolor[HTML]{FE0000}\textbf{}                        & \cellcolor[HTML]{FE0000}\textbf{}                        & \cellcolor[HTML]{FE0000}\textbf{} \\ \hline
\multicolumn{1}{|l|}{Deepseek reasoner} & \cellcolor[HTML]{C0C0C0}                                 & \cellcolor[HTML]{34FF34}\textbf{} & \cellcolor[HTML]{FE0000}\textbf{} & \cellcolor[HTML]{C0C0C0}          & \cellcolor[HTML]{C0C0C0}\textbf{}                        & \cellcolor[HTML]{34FF34}\textbf{}                        & \cellcolor[HTML]{FE0000}\textbf{}                        & \cellcolor[HTML]{C0C0C0}\textbf{}                        & \cellcolor[HTML]{C0C0C0}\textbf{}                        & \cellcolor[HTML]{C0C0C0}\textbf{}                        & \cellcolor[HTML]{FE0000}\textbf{}                        & \cellcolor[HTML]{C0C0C0}\textbf{} \\ \hline
\end{tabular}
\end{table}

Among the evaluated models, \textbf{Deepseek Chat} and \textbf{OpenAI 4.1 Nano} performed the worst, consistently returning a list of rules even when the input sentences contained clear ambiguities. These models failed to engage in clarification or exhibit caution when interpreting underspecified inputs.

In contrast, the \textbf{OpenAI o3} model demonstrated the most robust handling of ambiguity. It correctly identified unclear inputs in nearly all cases and either posed relevant follow-up questions or refrained from producing rules when sufficient information was lacking—an exception being sentence S2, where the model misinterpreted the constraint context.

Following o3, both \textbf{Gemini 2.5 Flash} and \textbf{Gemini 2.5 Pro} performed noticeably better than the remaining models, occasionally asking appropriate clarification questions. However, their behavior still fell short of the consistency and caution demonstrated by the o3 model, highlighting a substantial performance gap in ambiguity detection and interactive reasoning.

\subsection{Case Study}
\begin{sidewaysfigure}[ph!]
\begin{subfigure}{1\textwidth}
    \centering
\includegraphics[width=1\linewidth]{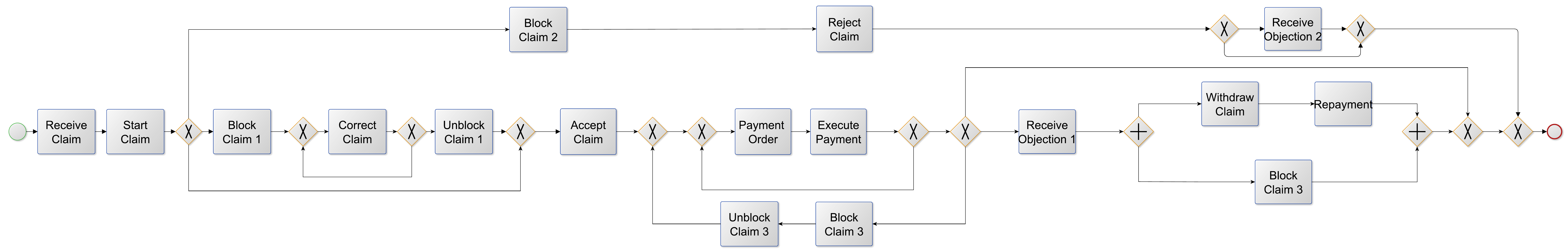}
    \caption{\small Normative model of the UWV claim handling process, extracted manually in collaboration with domain experts.}
    \label{normativ_UWV}
\end{subfigure}\\
\begin{subfigure}{1\textwidth}
    \centering
\includegraphics[width=1\linewidth]{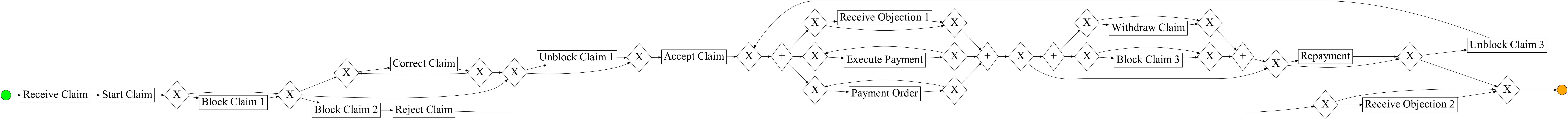}
    \caption{\small Model discovered by the IMr algorithm with parameter $sup=0.2$ and no rule used.}
    \label{fig:IMr_no_rule}
\end{subfigure}
\caption{Comparison between the normative process model and the automatically discovered model for the UWV claim handling process.}
\end{sidewaysfigure}

To evaluate the practical applicability of our approach in a real-world context, we conducted a case study in collaboration with the UWV, the Dutch employee insurance agency responsible for administering unemployment and disability benefits. The focus of this study is one of UWV's claim-handling processes. A normative process model, shown in Figure~\ref{normativ_UWV}, was developed jointly with domain experts who possess in-depth knowledge of the process. The corresponding event log comprises 144,046 cases, 16 distinct activities, and a total of 1,309,719 recorded events. 

To assess the usability of our proposed method, we conducted the case study in several stages. As an initial step, we asked a domain expert to describe the claim-handling process based on their knowledge, without being exposed to any discovered process model. The textual explanation provided by the expert is shown below and serves as the basis for constraint extraction. Given its strong performance in both generating high-quality rules and effectively handling ambiguities, as discussed earlier, the OpenAI o3 model was selected to extract declarative constraints from the expert’s input.

\begin{tcolorbox}[enhanced,breakable,colback=gray!5!white,colframe=blue!75!black,left=0.5mm, right=0.5mm]
\small
Our process is designed to handle claim applications within an insurance agency, where applicants request various types of benefits, and our system processes these applications through a structured workflow. There are three types of blocking mechanisms: Block Claim 1, Block Claim 2, and Block Claim 3, each with corresponding resolutions: Unblock Claim 1, Unblock Claim 2, and Unblock Claim 3. The process begins once a claim is received. Block Claim 1 occurs when initial issues are detected, leading to necessary corrections before the claim is unblocked and accepted, resulting in the client receiving between one to three payments. Block Claim 2 is applied when the claim is deemed ineligible for receiving any payments, resulting in rejection and potentially an objection from the client. Block Claim 3 is triggered if there is missing information, such as incorrect bank account details, after a claim is accepted; the client is notified to provide the correct details, and the claim is unblocked upon receipt of this information. Additionally, clients may file an objection after receiving payments, leading to the withdrawal of the claim and repayment of the received money to the agency.
\end{tcolorbox}

Following the process description, the LLM generated a set of follow-up questions aimed at clarifying ambiguities and resolving potential uncertainties in the input text.

 
\begin{tcolorbox}[enhanced,breakable,colback=gray!5!white,colframe=red!75!black,left=0.5mm, right=0.5mm]
\small
Before I can translate the description into precise constraints, could you please clarify the following points?

\begin{enumerate}
    \item The description mentions the activity \texttt{Unblock Claim 2}, but it is not present in the provided list of activities.
    \begin{itemize}
        \item Should \texttt{Unblock Claim 2} be added to the activity list, or does another listed activity cover this step?
    \end{itemize}

    \item Two objection activities exist in the list—\texttt{Receive Objection 1} and \texttt{Receive Objection 2}.
    \begin{itemize}
        \item Which objection is related to \texttt{Block Claim 2} (the ineligible-claim scenario), and which objection is related to the post-payment scenario that may lead to withdrawal and repayment?
    \end{itemize}

    \item For \texttt{Block Claim 1}, is the intended sequence always: 
    \texttt{Block Claim 1} $\rightarrow$ \texttt{Correct Claim} $\rightarrow$ \texttt{Unblock Claim 1} $\rightarrow$ \texttt{Accept Claim}?
    \begin{itemize}
        \item If other orders are possible, please specify.
    \end{itemize}

    \item After \texttt{Block Claim 3}, should \texttt{Unblock Claim 3} always occur (i.e., is it mandatory), or can a case end while still blocked?

    \item The statement ``the client receives between one to three payments'' suggests a numeric constraint on the number of \texttt{Execute Payment} occurrences.
    \begin{itemize}
        \item Is it acceptable to simplify this as ``at least one payment'' (since our available templates only allow \texttt{AtLeast1} and \texttt{AtMost1}), or do you want us to impose both \texttt{AtLeast1} and \texttt{AtMost1} (effectively limiting it to exactly one)?
    \end{itemize}

    \item Does every rejection following \texttt{Block Claim 2} require that no payment activities (\texttt{Payment Order} / \texttt{Execute Payment}) have occurred beforehand?
\end{enumerate}

With these clarifications, I can generate the precise declarative constraints in the required JSON format.
\end{tcolorbox}

The questions generated by the LLM demonstrate its strong ability to comprehend both the input text and the task at hand, enabling it to identify and address relevant ambiguities in a meaningful way. For instance, the domain expert's description contains certain inaccuracies, such as referencing an activity named \texttt{Unblock Claim 2}, which does not exist in the event log, and vague references to objection-related events without specifying whether \texttt{Receive Objection 1} or \texttt{Receive Objection 2} is meant. The OpenAI o3 reasoner model effectively detects these issues and formulates follow-up questions that are contextually appropriate and logically aligned with the input. After reviewing these questions with the domain expert, we collected the clarifying responses, which were then submitted to the LLM as the next message in the interaction.

\begin{tcolorbox}[enhanced,breakable,colback=gray!5!white,colframe=blue!75!black,left=0.5mm, right=0.5mm]
\small
\begin{enumerate}
    \item The previous description contained some inaccuracies. You are correct that \texttt{Unblock Claim 2} does not exist in the process. Please disregard any explanations referencing it.
    
    \item \texttt{Receive Objection 2} corresponds to the objection that may follow the second blocking mechanism (i.e., the ineligible-claim scenario), whereas \texttt{Receive Objection 1} is related to objections occurring after payment.
    
    \item If \texttt{Block Claim 1} occurs, the intended sequence you provided is generally correct. However, there may be multiple occurrences of \texttt{Correct Claim} before \texttt{Unblock Claim 1} is triggered.
    
    \item The occurrence of \texttt{Unblock Claim 3} after \texttt{Block Claim 3} is not mandatory. A case may end while still in a blocked state.
    
    \item Please assume that at least one payment must occur if the claim is accepted. This simplification is acceptable given the available constraint templates.
    
    \item Yes, in the case of rejection following \texttt{Block Claim 2}, no payment activities (i.e., \texttt{Payment Order} or \texttt{Execute Payment}) should have occurred beforehand.
\end{enumerate}

\end{tcolorbox}

The set of declarative constraints extracted after answering the questions is:

\begin{tcolorbox}[enhanced,breakable,colback=gray!5!white,colframe=red!75!black,left=0.5mm, right=0.5mm]
\small
\begin{itemize}
    \item Precedence(\texttt{Block Claim 1}, \texttt{Correct Claim})
    \item Response(\texttt{Block Claim 1}, \texttt{Correct Claim})
    \item Precedence(\texttt{Correct Claim}, \texttt{Unblock Claim 1})
    \item Response(\texttt{Correct Claim}, \texttt{Unblock Claim 1})
    \item Precedence(\texttt{Block Claim 1}, \texttt{Unblock Claim 1})
    \item Response(\texttt{Unblock Claim 1}, \texttt{Accept Claim})
    \item Response(\texttt{Accept Claim}, \texttt{Execute Payment})
    \item Precedence(\texttt{Execute Payment}, \texttt{Receive Objection 1})
    \item Response(\texttt{Receive Objection 1}, \texttt{Withdraw Claim})
    \item Precedence(\texttt{Withdraw Claim}, \texttt{Repayment})
    \item Response(\texttt{Withdraw Claim}, \texttt{Repayment})
    \item Response(\texttt{Block Claim 2}, \texttt{Reject Claim})
    \item Precedence(\texttt{Block Claim 2}, \texttt{Reject Claim})
    \item NotSuccession(\texttt{Payment Order}, \texttt{Reject Claim})
    \item NotSuccession(\texttt{Execute Payment}, \texttt{Reject Claim})
    \item Precedence(\texttt{Accept Claim}, \texttt{Block Claim 3})
    \item Precedence(\texttt{Block Claim 3}, \texttt{Unblock Claim 3})
    \item NotCoExistence(\texttt{Accept Claim}, \texttt{Reject Claim})
\end{itemize}

\end{tcolorbox}

To establish a reference model using the core process discovery technique, we applied the IMr discovery framework with a minimum support threshold of $sup = 0.2$, without incorporating any domain rules. The resulting model is shown in Fig.~\ref{fig:IMr_no_rule}. This model exhibits several issues when compared to the normative model. For example:

\begin{itemize}
    \item \texttt{Block Claim 1} should only be applicable to claims that are ultimately accepted. However, the model permits its use for both accepted and rejected claims, which is inconsistent with the intended behavior.
    
    \item The logical ordering of \texttt{Payment Order}, \texttt{Execute Order}, and \texttt{Receive Objection 1} is incorrect. All executed payments must first be ordered, and it is not permissible to receive an objection before any payment has been executed.
    
    \item The placement of the \texttt{Unblock Claim 3} transition within a loop that allows repeated repayments and claim withdrawals is problematic. This activity is intended to initiate the payment cycle after a claim has been blocked using the third mechanism, specifically when the issue has been resolved.
\end{itemize}

\begin{sidewaysfigure}[ph!]
\begin{subfigure}{1\textwidth}
    \centering
\includegraphics[width=1\linewidth]{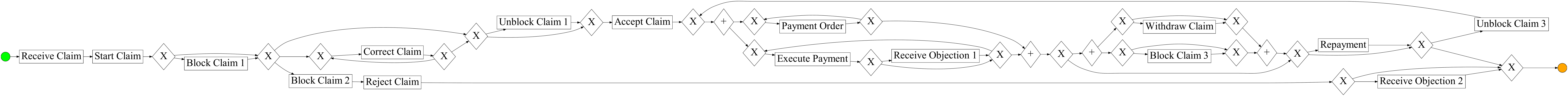}
    \caption{\small Process model discovered by IMr with \( \mathit{sup} = 0.2 \), incorporating rules extracted from the initial process description and follow-up Q\&A interactions.}
    \label{fig:uwv1}
\end{subfigure}\\
\begin{subfigure}{1\textwidth}
    \centering
\includegraphics[width=1\linewidth]{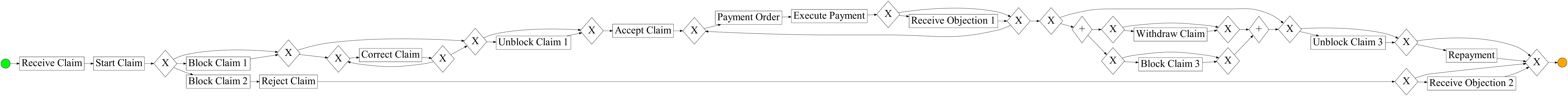}
    \caption{\small Model discovered by IMr with $sup=0.2$ considering the rules extracted after the first feedback provided by the domain expert.}
    \label{fig:uwv2}
\end{subfigure}\\
\begin{subfigure}{1\textwidth}
    \centering
\includegraphics[width=1\linewidth]{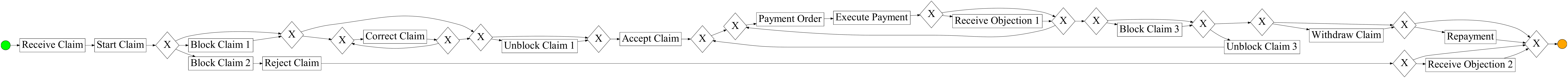}
    \caption{\small Model discovered by IMr with $sup=0.2$ considering the rules extracted after the second feedback provided by the domain expert.}
    \label{fig:uwv3}
\end{subfigure}
\caption{Incremental refinement of the discovered process model for the UWV claim handling case using IMr with $sup=0.2$.}
\end{sidewaysfigure}

Figure~\ref{fig:uwv1} illustrates the process model discovered using the same event log in combination with the extracted declarative rules. A comparison with both the normative model and the model discovered without rules reveals that incorporating business context prior to discovery has had limited impact on improving model quality. The resulting model continues to exhibit similar issues. This is primarily because the provided contextual information focuses on process aspects that are either already discoverable without additional rules or cannot be effectively enforced due to representational biases inherent in the IMr algorithm.

As the next stage of the case study, the discovered process model was presented to the domain expert for evaluation. The expert was asked to identify potential quality issues and deviations from the expected process behavior. The resulting feedback, formulated as a textual description, was then provided to the LLM. The goal was to extract a set of declarative rules from this expert feedback that could guide the process discovery algorithm towards models that better reflect the underlying domain knowledge.

\begin{tcolorbox}[enhanced,breakable,colback=gray!5!white,colframe=blue!75!black,left=0.5mm, right=0.5mm]
\small
The first block claim mechanism must never be triggered for rejected claims. Block Claim 1 applies only to accepted cases, while Block Claim 2 is designated exclusively for rejected ones. Additionally, the withdrawal of a claim and the repayment activity must not occur more than once. However, the loop initiated by Unblock Claim 3 currently allows multiple occurrences of these actions, which is undesirable. Furthermore, a payment execution must always follow a payment order. It is invalid for the first receipt of an objection to occur before any payment order has been issued.
 \end{tcolorbox}

After the consideration of the domain expert input, these constraints are extracted by the LLM:

\begin{tcolorbox}[enhanced,breakable,colback=gray!5!white,colframe=red!75!black,left=0.5mm, right=0.5mm]
\small
\begin{itemize}
    \item Precedence(\texttt{Accept Claim}, \texttt{Block Claim 3})
    \item Response(\texttt{Accept Claim}, \texttt{Execute Payment})
    \item NotCoExistence(\texttt{Accept Claim}, \texttt{Reject Claim})
     \item Response(\texttt{Block Claim 1}, \texttt{Correct Claim})
    \item Precedence(\texttt{Block Claim 1}, \texttt{Correct Claim})
    \item Precedence(\texttt{Block Claim 1}, \texttt{Unblock Claim 1})
    \item  Response(\texttt{Block Claim 2}, \texttt{Reject Claim})
    \item Precedence(\texttt{Block Claim 2}, \texttt{Reject Claim})
    \item Precedence(\texttt{Block Claim 3}, \texttt{Unblock Claim 3})
    \item Response(\texttt{Correct Claim}, \texttt{Unblock Claim 1})
    \item Precedence(\texttt{Correct Claim}, \texttt{Unblock Claim 1})
    \item Precedence(\texttt{Execute Payment}, \texttt{Receive Objection 1})
    \item NotSuccession(\texttt{Execute Payment}, \texttt{Reject Claim})
    \item NotSuccession(\texttt{Payment Order}, \texttt{Reject Claim})
     \item Response(\texttt{Receive Objection 1}, \texttt{Withdraw Claim})
    \item Response(\texttt{Unblock Claim 1}, \texttt{Accept Claim})
    \item Response(\texttt{Withdraw Claim}, \texttt{Repayment})
    \item Precedence(\texttt{Withdraw Claim}, \texttt{Repayment})
    \item \colorbox{green!20}{NotCoExistence(\texttt{Block Claim 1}, \texttt{Reject Claim})}
    \item \colorbox{green!20}{NotCoExistence(\texttt{Block Claim 2}, \texttt{Accept Claim})}
    \item \colorbox{green!20}{AtMost1(\texttt{Repayment})}
    \item \colorbox{green!20}{AtMost1(\texttt{Withdraw Claim})}
    \item \colorbox{green!20}{Precedence(\texttt{Payment Order}, \texttt{Execute Payment})}
    \item \colorbox{green!20}{Precedence(\texttt{Payment Order}, \texttt{Receive Objection 1})}
\end{itemize}
\end{tcolorbox}

The rules highlighted in green represent newly added constraints that were not part of the previous rule set.
 The process model shown in Fig.~\ref{fig:uwv2} was discovered by integrating the extracted set of constraints with the input event log using the IMr framework. This revised model was then presented to the domain expert for evaluation. The expert confirmed that, in comparison to the previous model, the new one more accurately reflects the actual process, which is further supported by its improved alignment with the normative model depicted in Fig.~\ref{normativ_UWV}. However, the expert also identified one major issue that still needs to be addressed. This remaining shortcoming is described in detail below.

\begin{tcolorbox}[enhanced,breakable,colback=gray!5!white,colframe=blue!75!black,left=0.5mm, right=0.5mm]
\small
Every occurrence of Unblock Claim 3 must be followed by at least one payment activity.
 \end{tcolorbox}

The following constraint was identified as an additional refinement to complement the previously extracted set of constraints.

 \begin{tcolorbox}[enhanced,breakable,colback=gray!5!white,colframe=red!75!black,left=0.5mm, right=0.5mm]
\small
\begin{itemize}
\item ...
    \item \colorbox{green!20}{Response(\texttt{Unblock Claim 3}, \texttt{Execute Payment})}
\end{itemize}
\end{tcolorbox}

The inclusion of this single constraint, together with the previously extracted rules and the event log, resulted in the discovery of the process model shown in Figure~\ref{fig:uwv3}, which demonstrates improved alignment with the domain expert’s knowledge. 

Some of the rules asserted by the domain expert, and correctly identified by the LLM, are still not fully satisfied by the discovered model. This is due to inherent limitations of the IMr technique, particularly its representational bias. For instance, all discovered models permit traces in which \texttt{Block Claim 1} is skipped, yet \texttt{Unblock Claim 1} occurs. As a result, the constraint Precedence(\texttt{Block Claim 1}, \texttt{Unblock Claim 1}) is violated. The recursive structure of the IMr algorithm can lead to the omission of such long-term dependencies, especially when they span across multiple recursive steps.

\section{Conclusion}
\label{sec:conclusion}
Event logs extracted from information systems are not the only source of information relevant to process discovery. Incorporating domain knowledge remains an underexplored yet essential dimension in process mining, particularly for producing models that align more closely with real-world operations. This paper proposed a novel framework that leverages LLMs to extract declarative constraints from natural language descriptions provided by domain experts. These constraints serve as supplementary input to guide the discovery of process models that are not only behaviorally sound but also better aligned with expert understanding and organizational context.

We conducted a comprehensive evaluation of the framework, assessing its performance across multiple configurations and LLMs. These included variations in prompt design (e.g., with or without examples), input granularity (sentence-level vs. paragraph-level prompting), and the LLM’s handling of ambiguous inputs. The results demonstrate the feasibility and potential of using LLMs for this task, showing promising accuracy in rule extraction and robustness in dealing with incomplete or ambiguous input. Furthermore, a case study using a real-life event log from UWV, complemented by direct interaction with domain experts, demonstrated the practical usability of the framework in real-world settings.

Our framework introduces several key innovations: it enables the encoding of domain knowledge into formal rules, supports iterative and interactive refinement through expert feedback, and integrates seamlessly with the IMr discovery algorithm to balance data-driven learning with knowledge-guided modeling. 
Ultimately, this work lays the foundation for a more human-in-the-loop paradigm in process discovery, where domain expertise and automated analysis of event data complement one another to produce interpretable, and accurate process models.

\bibliographystyle{splncs04}
\bibliography{lit_new}






\end{document}